%% file: MAST_ICLR2027.tex
\documentclass{article}
\usepackage{iclr2027_conference,times}

\usepackage{graphicx}
\usepackage{algorithm}
\usepackage{algorithmic}
\usepackage{amsmath}
\usepackage{amssymb}
\usepackage{multirow}
\usepackage{booktabs}
\usepackage{capt-of}
\usepackage{hyperref}
\hypersetup{hidelinks}
\usepackage{url}

\newcommand{\FigFivePanelWidth}{0.47\linewidth}
\newcommand{\PopeTablePanelWidth}{0.51\linewidth}
\newcommand{\AblationTablePanelWidth}{0.52\linewidth}
\newcommand{\SubspaceTablePanelWidth}{0.44\linewidth}

\title{MAST: Medoid-Anchored Subspace Steering for Visual Grounding in Multimodal LLMs}
\author{Zhitao Chen$^{1}$ \quad Zhenxin Cai$^{1}$ \quad Jingyuan Li$^{2}$ \quad Jin Sha$^{1}$\\[0.6ex]
{\normalfont\small $^{1}$School of Electronic Science and Engineering, Nanjing University, China}\\
{\normalfont\small $^{2}$College of Electronic Science and Technology, National University of Defense Technology, China}}

\begin{document}

\maketitle

\begin{abstract}
Existing activation steering methods for multimodal large language models (MLLMs) use the global activation mean as the steering vector, which tends to cause cross-task interference and unstable performance. We trace this failure to the multi-centered geometry of task activations. The global mean lands in the empty region between clusters and is faithful to no single task. Moreover, the steering signal in each task-pure local cluster is low-rank, and different query tasks benefit from emphasizing different families of local-basis directions under the shared routing protocol. To address this, we propose \textbf{MAST} (\textbf{M}edoid-\textbf{A}nchored \textbf{S}ubspace S\textbf{t}eering), a training-free, input-adaptive activation steering framework. MAST groups samples by task, performs multi-center clustering within each group, and selects one representative medoid per cluster. At inference time, each query is routed to multiple representative medoids. For each matched cluster, a low-rank local steering direction is extracted and aggregated across clusters to produce the final steering vector. Building on MAST, \textbf{MAST-PR} (Probe-Reweighted) uses a lightweight task-aware probe to reweight task-specialized local-direction families, yielding a further average improvement of +8.09 points over MAST.
\end{abstract}


\section{Introduction}

\begin{figure}[htbp]
    \centering
    \includegraphics[width=0.48\linewidth]{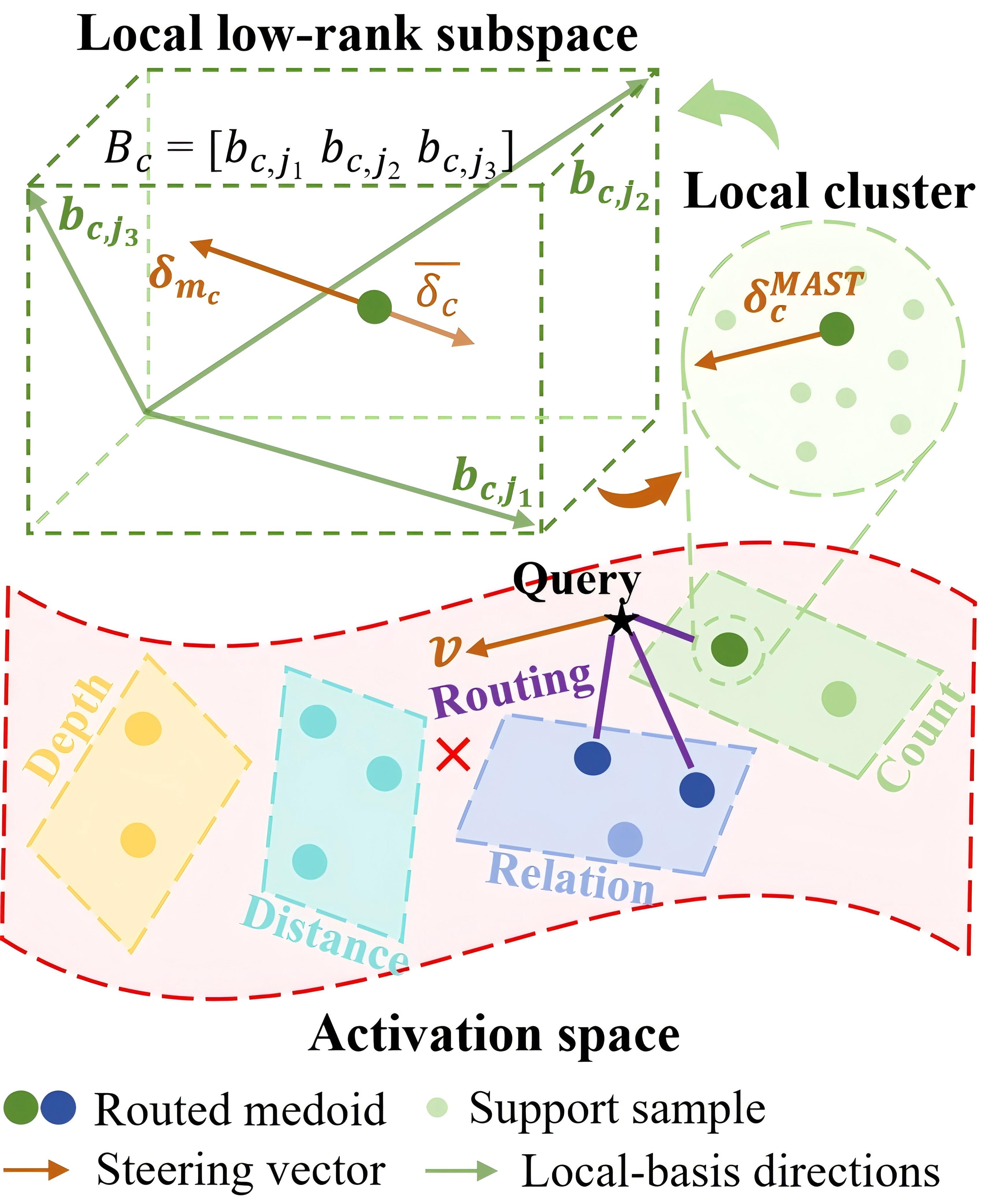}
    \caption{An illustration of the MAST principle. Visual tasks form their own
    local clusters in the activation space, and the global mean (the red cross
    $\times$) lands in the empty region between clusters. MAST
    routes a query (the $\star$) to the
    nearest medoid and, within its local cluster, represents the local steering
    signal with low-rank local subspace directions. MAST-PR can reweight these
    directions using task-specialized weights.}
    \label{fig:figure1}
\end{figure}

By modifying a model's internal activations during the forward pass, activation steering has emerged as a training-free, inexpensive approach for controlling language model behavior~\citep{feng2026finegrained,chalnev2024improvingsteeringvectorstargeting}. Recent work extends it to multimodal large language models (MLLMs), computing the mean of activation differences from paired positive and negative examples and injecting it as a single steering vector into the residual stream, yielding substantial gains on visual tasks~\citep{phute2025visorvisualinputbasedsteering}.

However, current gains are highly uneven. \citet{gan2025textualsteeringvectorsimprove} report clear improvements on spatial relation but only marginal gains elsewhere, indicating that the global-mean direction cannot serve different visual tasks and succeeds only when the target behavior concentrates around the global mean. The cause is that different visual operations occupy separate activation regions, each forming its own cluster, so the global mean lands in the empty region between clusters, unable to guide any single task precisely (Figure~\ref{fig:figure1}).

The deeper reason is a conflict among local steering directions across clusters that worsens with the visual discrepancy between tasks, as opposing directions cancel under the coarse global mean and are buried in noise~\citep{rodriguez2025controlling}. Within every task-pure cluster, meanwhile, the steering signal concentrates in a low-rank subspace. These observations play distinct roles. Routing avoids averaging conflicting clusters, while the local basis provides a compact set of directions whose contributions can be reweighted for the query task. We call such a task-preferred, cross-cluster family of corresponding local-basis directions \emph{task-specialized}; this is an operational performance notion, not a claim that one task-pure cluster contains multiple task identities. The global mean loses both locality and this fine-grained control.

Motivated by this, we propose \textbf{MAST} (\textbf{M}edoid-\textbf{A}nchored \textbf{S}ubspace S\textbf{t}eering), a training-free, input-adaptive framework. MAST groups support samples by visual task, clusters within each group, and selects each cluster's medoid as its representative. At inference each query is routed to its nearest few representatives, and a low-rank local direction extracted from each matched cluster is aggregated into the final steering vector, so every query receives a faithful direction on its own local manifold. Building on it, \textbf{MAST-PR} (Probe-Reweighted) uses a lightweight task-aware probe to reweight the task-specialized local-direction families, so the directions empirically preferred by the query task receive greater emphasis.

On LLaVA-v1.5-7B (hereafter LLaVA-v1.5), MAST substantially outperforms all baselines on the CV-Bench visual understanding benchmark~\citep{tong2024cambrian}, improving spatial relation, counting, distance, and depth estimation by +19.42, +6.35, +16.25, and +26.25 points respectively, with consistent gains on the POPE (Polling-based Object Probing Evaluation) hallucination benchmark~\citep{li2023pope}. MAST-PR attains a further +8.09 points on average. Our main contributions are as follows.

\begin{itemize}
\item We characterize visual-task activations in MLLMs as a multi-center distribution with conflicting local directions, while each task-pure cluster contains a low-rank steering subspace. This explains why global-mean steering fails on heterogeneous visual tasks and supplies compact local directions for task-specialized reweighting.

\item We propose MAST, which realizes input-adaptive steering through multi-center medoid routing and local low-rank subspace extraction, setting new state-of-the-art results on CV-Bench and POPE. Controlled ablation identifies input-adaptive multi-center prototype routing as the primary driver of equal-weight MAST, while local subspaces enable task-specialized reweighting in MAST-PR.

\item We propose MAST-PR, which reweights each local subspace's directions per task and improves MAST by +8.09 points. Controlled task-oracle comparisons verify the utility of task-specialized local-direction families.
\end{itemize}

\section{Related Work}

Early methods build a direction from the activation differences of a few contrastive samples, enough to modulate the sentiment, topic, factuality, or refusal of generated text~(\mbox{\citealp{turner2024steeringlanguagemodelsactivation}}; \mbox{\citealp{rimsky-etal-2024-steering}}; \mbox{\citealp{stolfo2025improving}}; \mbox{\citealp{zhang2025differentiateddirectionalinterventionframework}}), and \citet{chalnev2024improvingsteeringvectorstargeting} use a sparse autoencoder to quantify its causal effect per behavior. All control an entire behavior with a single input-independent direction, implicitly assuming the target activations form a single center.

This has been extended to MLLMs~\citep{10.1609/aaai.v40i41.40773, 10.1609/aaai.v40i40.40720}, by transferring a mean-shift direction from a text-only backbone~\citep{gan2025textualsteeringvectorsimprove}, compensating for fine-tuning-induced concept drift~\citep{11445319}, or training a module that predicts a steering vector per input~\citep{parekh2025neurips-learning}. These either stay confined to a fixed direction or trade training for adaptivity, and all apply the vector to the activation as a whole, without distinguishing which components drive the target behavior.

For finer control, some works intervene on only a few discriminative dimensions~\citep{feng2026finegrained} or use a probe to set the injection strength adaptively~\citep{10.1145/3696410.3714640, huang2026ampsadaptivemodalitypreference, kang2026enhancinginstructionfollowingllms, valentino2026mitigatingcontenteffectsreasoning}, achieving strength adaptivity but treating each direction as isolated, overlooking the geometric structure the directions jointly form.

A recent line examines this geometric structure~\citep{jang2026tvsstesttimevisualsubspace}, estimating one set of directions over the whole representation space via principal component analysis~\citep{ICLR2025_b4008025} or QR orthogonalization~\citep{li2026suppressingpriorcomparisonhallucinationsradiology} and applying it to all inputs indiscriminately. \citet{kayan2025prototypebaseddynamicsteeringlarge} route each input to prototypes clustered from contrastive activations, but form only a single prototype-weighted direction. MAST goes further by identifying multiple task-specialized centers and extracting a low-rank local direction from each center's neighborhood; MAST-PR then applies task-specialized reweighting over the resulting cross-cluster direction families, so directions for different tasks and inputs do not interfere.

\section{Heterogeneous Activation Geometry in Visual Tasks}
\label{sec:motivation}

We first characterize the geometry of visual-task activations to explain why the global mean fails. All three observations use layer~16 of LLaVA-v1.5 on the CV-Bench support samples, with label-grouped $K$-medoids clustering~\citep{PARK20093336} ($K{=}5$) within each task.

\begin{figure}[htbp]
    \centering
    \begin{minipage}[t]{0.54\linewidth}
        \centering
        \vspace{0pt}
        \includegraphics[width=0.98\linewidth]{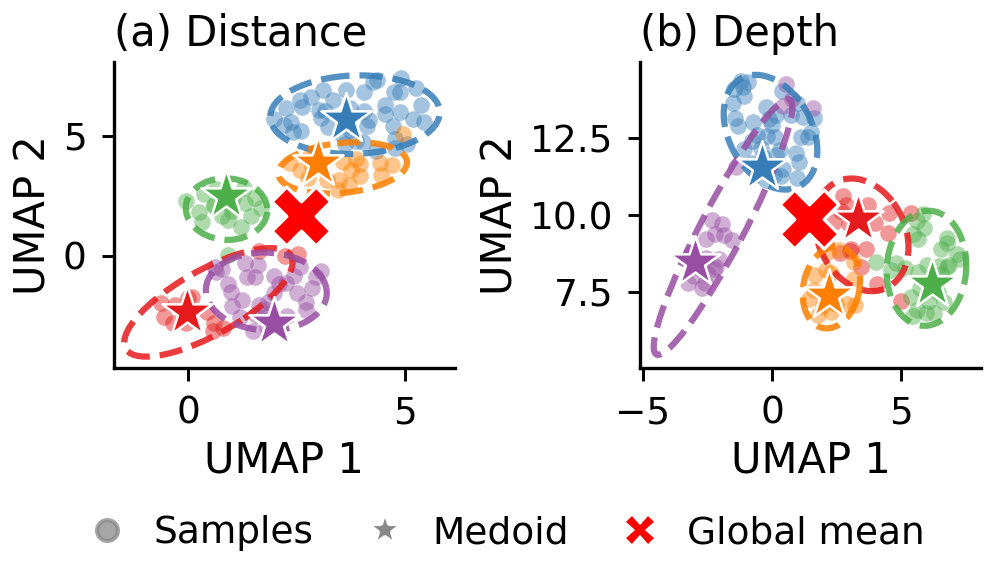}
        \caption{UMAP (Uniform Manifold Approximation and Projection)~\citep{mcinnes2020umapuniformmanifoldapproximation} projection of CV-Bench support activations (Distance and
        Depth). Samples colored by cluster, stars mark medoids, black cross the
        global mean. Full four-task version in
        Appendix~\ref{sec:appendix_geometry}.}
        \label{fig:obs1_pca_scatter}
    \end{minipage}\hfill
    \begin{minipage}[t]{0.43\linewidth}
        \centering
        \vspace{0pt}
        \includegraphics[width=0.86\linewidth]{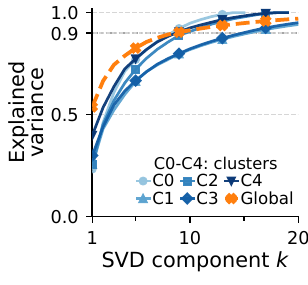}
        \caption{Cumulative variance of the intra-cluster steering vectors under SVD
        (Distance). Solid lines are per-cluster curves, black dashed the
        no-clustering baseline, dotted grey the 90\% threshold. Full four-task
        version in Appendix~\ref{sec:appendix_geometry}.}
        \label{fig:singular_value_cumvar}
    \end{minipage}
\end{figure}

Visual-task activations form multiple local clusters rather than a single center. Under UMAP the global mean lands in the empty region between clusters, faithful to none (Figure~\ref{fig:obs1_pca_scatter}). Routing test queries to their nearest medoid, the combined 20-cluster setting over the four tasks reaches 76.69\% accuracy, so the cluster layout already encodes task identity. Full metrics and four-task projections are in Appendix~\ref{sec:appendix_geometry}.

The steering signal within each task-pure cluster concentrates in an extremely low-rank subspace. Under singular value decomposition (SVD) of each cluster's sample steering vectors (Figure~\ref{fig:singular_value_cumvar}), across all clusters of all four tasks only a handful of leading singular directions capture at least 90\% of the variance, versus an activation dimension of 4096, so a local direction truncates cleanly almost without loss. Without clustering, the leading direction is merely the average of the clusters' conflicting directions, nearly orthogonal to each and faithful to none. Crucially, the local-basis directions are not interchangeable under the shared routing protocol. Our direction-specialization scan (Appendix~\ref{sec:appendix_direction}) shows that different tasks benefit most from emphasizing different same-rank direction families across the routed clusters (on LLaVA-v1.5, the second, fifth, and sixth for Distance, Count, and Relation). Thus, the low-rank basis supports task-specialized reweighting without implying that a single task-pure cluster contains multiple tasks. Full curves and per-task rank counts are in Appendix~\ref{sec:appendix_geometry}.

The steering directions of different clusters conflict, often nearly orthogonal or opposite. The average pairwise cosine similarity between the five medoid directions is lowest for the geometric tasks (0.030 for Distance and 0.055 for Depth, i.e.\ almost orthogonal), intermediate for Relation (0.223, with one cluster negatively correlated with all others), and highest for Count (0.356, yet with an outlier cluster that pulls the global mean off course). Averaging nearly orthogonal directions yields a global mean that shrinks and deviates from every local center, losing steering capacity. This is why global-mean methods degrade sharply on Depth and Distance and MAST gains most there (26.25 and 16.25 points on Depth and Distance, with full heatmaps and per-cluster values in Appendix~\ref{sec:appendix_geometry}).

Together these observations dictate MAST's design. MAST routes each query to its cluster, extracts a clean local direction from the low-rank intra-cluster signal, and avoids averaging across conflicting clusters. The task-dependent utility of the local direction families in turn motivates the per-task reweighting of MAST-PR.

\section{Method}
\label{sec:method}

\begin{figure}[htbp]
    \centering
    \includegraphics[width=0.92\linewidth]{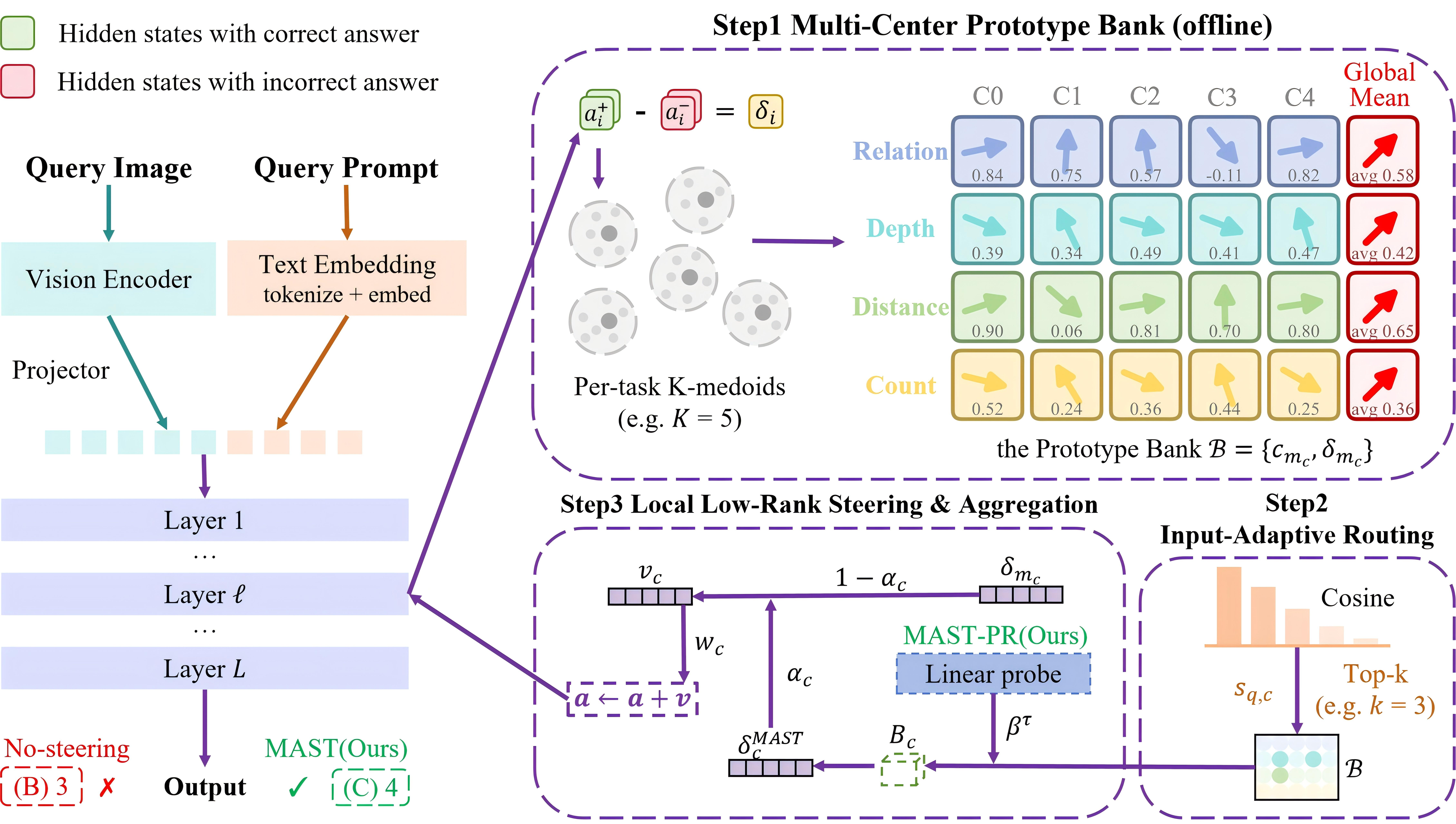}
    \caption{Overview of \emph{MAST}. \emph{Step 1 (offline).} Within each
    task, $K$-medoids clustering ($K{=}5$) builds the prototype bank
    $\mathcal{B}$ of $20$ medoids, whose directions are largely misaligned with
    the task global mean. \emph{Step 2 (online).} Adaptive routing by
    $s_{q,c}$ selects the top-$k$ medoids ($k{=}3$). \emph{Step 3 (online).}
    Each selected cluster is blended into $v_c$ and summed into the steering
    vector $v$. Contrastive directions are extracted from teacher-forced
    support completions, whereas routing uses separately extracted answer-free
    prompt contexts (Section~\ref{sec:prelim}). The extension \emph{MAST-PR}
    reweights the task-specialized directions $\beta^{\tau}$ via a lightweight
    probe.}
    \label{fig:overview}
\end{figure}

The activation geometry of the previous section dictates MAST's design, which first constructs multi-center prototypes, then extracts intra-cluster low-rank directions, and finally aggregates them adaptively per input at inference (Figure~\ref{fig:overview}).

\subsection{Preliminaries}
\label{sec:prelim}

We intervene at a single fixed layer, adding a vector $v$ to the residual-stream activation $a\in\mathbb{R}^{4096}$, i.e.\ $a\leftarrow a+v$. For each support sample $i$, let $P_i$ denote the image--question prompt. We teacher-force a correct completion $y_i^{+}$ and an incorrect completion $y_i^{-}$ in two separate forward passes. The positive and negative activations $a_i^{+}$ and $a_i^{-}$ are the mean layer-$\ell$ residual-stream states over their respective assistant-completion token spans (including a terminal chat-format token when present),
\begin{equation}
\delta_i = a_i^{+} - a_i^{-}
\end{equation}
giving the single-sample contrastive direction toward the target behavior. Routing uses a separate answer-free prompt context. Here, $c_i$ is the layer-$\ell$ residual state at the last prompt token immediately before the assistant completion in $P_i$, and $c_q$ is extracted at the same position from the answer-free query prompt $P_q$. Thus, neither $c_i$ nor $c_q$ contains a gold or candidate answer. The support set is partitioned by task into $\mathcal{S}_\tau=\{(c_i,\delta_i)\}$, and global-mean methods take $v=\frac{1}{N}\sum_i\delta_i$ for all queries. The exact support/query sequence is given in Appendix~\ref{sec:appendix_method}.

\subsection{Multi-Center Prototype Construction}
\label{sec:prototype}

Since samples of the same task scatter across separated clusters with different directions, the global mean averages these into one vector. MAST instead selects within each task a few real representative samples, each serving only the queries near it, and using a real sample rather than a centroid keeps the representative on the data manifold.

Concretely, for task $\tau$ we run $K$-medoids on its answer-free context set $\{c_i: i\in\mathcal{S}_\tau\}$, obtaining $K$ non-overlapping clusters whose medoid is the member with smallest sum of squared distances to the others,
\begin{equation}
m_c=\arg\min_{i\in c}\sum_{j\in c}\lVert c_i-c_j\rVert^{2}
\end{equation}
We store each medoid's context $c_{m_c}$ (for routing) and steering vector $\delta_{m_c}$ in a prototype bank $\mathcal{B}$, built offline with no gradient updates and no test labels.

\subsection{Local Low-Rank Steering and Adaptive Aggregation}
\label{sec:steering}

Since the intra-cluster steering vectors concentrate in a very low-rank subspace, we center them about the cluster mean $\bar\delta_c=\frac{1}{|c|}\sum_{i\in c}\delta_i$ and apply SVD, taking the leading $r$ right singular vectors as a local orthonormal basis $B_c$. For a direction-weight vector $\gamma\in\mathbb{R}^r$, define the weighted local reconstruction
\begin{equation}
\tilde\delta_c(\gamma)=\bar\delta_c+\sum_{j=1}^{r}\gamma_j\,\langle\delta_{m_c}-\bar\delta_c,b_{c,j}\rangle b_{c,j}.
\end{equation}
Base MAST uses equal weights, $\delta_c^{\mathrm{MAST}}=\tilde\delta_c(\mathbf{1}_r)=\bar\delta_c+B_cB_c^{\top}(\delta_{m_c}-\bar\delta_c)$, which can be precomputed offline. In base MAST, this reconstruction is a local refinement; multi-center routing drives the main gain, while the basis enables MAST-PR reweighting. MAST-PR instead computes query-dependent weights and reconstructs its local direction online.

At inference, given an answer-free query context $c_q$, we compute $s_{q,c}=\cos(c_q, c_{m_c})$ against all medoid contexts and select the $k$ most similar clusters, with index set $\mathcal{N}_k(q)$. These similarities are normalized into cross-cluster aggregation weights
\begin{equation}
w_c=\frac{\exp(s_{q,c}/T)}{\sum_{c'\in\mathcal{N}_k(q)}\exp(s_{q,c'}/T)}
\end{equation}
For each selected cluster we blend its raw medoid direction $\delta_{m_c}$ with $\delta_c^{\mathrm{MAST}}$ through a coefficient $\alpha_c\in[\alpha_{\min},\alpha_{\max}]=[0.05,0.80]$ that decreases with the cluster's local reliability $\rho_c\in[0,1]$, a fixed convex combination of relative cluster size, compactness, and routing confidence (full formulas in Appendix~\ref{sec:appendix_method}),
\begin{equation}
v_c=(1-\alpha_c)\,\delta_{m_c}+\alpha_c\,\delta_c^{\mathrm{MAST}}
\end{equation}
When reliability is low, $\alpha_c$ is large and the direction leans on the denoised reconstruction. When high, more of the medoid's raw signal is kept. The final steering vector is the similarity-weighted sum over the $k$ clusters,
\begin{equation}
v=\sum_{c\in\mathcal{N}_k(q)} w_c\, v_c
\end{equation}
At test time the query contains only the image and question, with no gold or candidate answer. We install $v$ as a layer-$\ell$ residual-stream hook and add it to all merged prompt positions in the initial prompt forward pass and to each subsequently generated token. Each query thereby receives a direction on its own local manifold rather than a globally aliased mean. Algorithm~\ref{alg:mast} summarizes the procedure.

\subsection{MAST-PR: Task-Specialized Local-Subspace Reweighting}
\label{sec:mastpr}

MAST-PR reweights the same-rank local-basis families across routed clusters using a lightweight four-way logistic-regression probe on the answer-free context $c_q$. Task-specific ranks and weights are selected only on validation data, then softly mixed as $\beta_{q,c,j}=\sum_{\tau}p(\tau\mid c_q)\beta_{c,j}^{\tau}$ to reconstruct
\begin{equation}
\delta_{q,c}^{\mathrm{PR}}=\bar\delta_c+\sum_{j=1}^{r}\beta_{q,c,j}\,\langle\delta_{m_c}-\bar\delta_c,\,b_{c,j}\rangle\,b_{c,j}.
\end{equation}
MAST-PR substitutes $\delta_{q,c}^{\mathrm{PR}}$ for $\delta_c^{\mathrm{MAST}}$ in the unchanged blending and aggregation pipeline; MAST is the all-ones special case. It improves MAST by $8.09$ points on average. Its main limitation is Depth. The probe assigns $36.50\%$ mean probability to Depth and $49.62\%$ to Distance, associated with a drop from $99.58\%$ to $90.42\%$; this affects only the optional reweighting extension, not MAST routing. Under uniform posterior noise, MAST-PR retains a positive gain through $\epsilon{=}0.75$; a fixed confidence gate recovers Depth to the equal-weight MAST level while retaining a $5.83$-point average gain. Appendix~\ref{sec:appendix_mastpr} gives the complete weighting rule and direction-family definition, while Appendix~\ref{sec:appendix_router} reports probe diagnostics, posterior perturbations, the causal Depth-to-Distance intervention, and the gate.

\section{Experiments}
\label{sec:exp}

\subsection{Experimental Setup}
\label{sec:setup}

Our primary model is LLaVA-v1.5~\citep{10655294}, with cross-model validation on Qwen2-VL-7B~\citep{wang2024qwen2vlenhancingvisionlanguagemodels} (hereafter Qwen2-VL). We evaluate on CV-Bench (four visual grounding subtasks, Relation, Count, Distance, Depth) and POPE (three COCO-based (Common Objects in Context)~\citep{10.1007/978-3-319-10602-1_48} splits, Random, Popular, Adversarial), assessing visual grounding and object hallucination. To avoid tuning on the evaluation data, we split the official CV-Bench test set task-stratified into a validation split ($20\%$, for all model selection) and a disjoint test split ($80\%$, reported), and fix each model's intervention layer by its own layer scan (layer~16 for LLaVA-v1.5, layer~4 for Qwen2-VL, Section~\ref{sec:layer}). MAST uses one default hyperparameter set throughout, never tuned per method, whereas we grant every steering baseline the more favorable treatment of separately searching its best injection strength on the validation split. Dataset sizes, hyperparameters, the baseline tuning protocol, and implementation details are in Appendix~\ref{sec:appendix_setup}, and our code is provided in the supplementary material.

As a validation-only check of the completion-to-generation transfer assumption, we additionally evaluate an answer-free instruction-contrast control; under the frozen MAST configuration it does not retain the main protocol's benefit, and its complete construction and results are reported in Appendix~\ref{sec:appendix_causal_control}.

\textbf{POPE protocol.} For POPE, we construct one shared prototype bank from all $1{,}200$ contrastive support examples in the POPE training split ($400$ from each of the Random, Popular, and Adversarial subsets). We group the support examples by target object and run $K{=}5$ medoids within each object group; this object-labeled bank is shared by all three POPE test splits. No CV-Bench support example or CV-Bench task label is used to construct the bank. Support images are drawn from COCO train2014, whereas POPE test images are drawn from COCO val2014. We use the fixed MAST settings above, without a POPE validation search.

\subsection{Main Results}
\label{sec:main}

We report CV-Bench results in Table~\ref{tab:main_cvbench}, then POPE. The baselines cover four steering paradigms, global contrastive directions (CAA, Act-Add), a global low-rank subspace (RepE), adaptive single-center methods (Linear-AcT, L2S), and multi-center cluster-routed steering (ACT and our MAST).

\begin{table}[htbp]
    \centering
    {\small
    \setlength{\tabcolsep}{2pt}
    \begin{tabular}{lrrrrr rrrrr}
        \hline
        & \multicolumn{5}{c}{LLaVA-v1.5} & \multicolumn{5}{c}{Qwen2-VL} \\
        \cline{2-6}\cline{7-11}
        Method & Relation & Count & Distance & Depth & Average & Relation & Count & Distance & Depth & Average \\
        \hline
        No-steering        & 59.62 & 47.46 & 23.33 & 73.33 & 50.94 & 78.08 & 67.14 & 77.71 & 86.67 & 77.40 \\
        Norm-Rnd           & 59.42 & 49.52 & 27.08 & 73.33 & 52.34 & 77.31 & 66.51 & 77.29 & 87.08 & 77.05 \\
        CAA                & 56.15 & 46.67 & 22.92 & 73.12 & 49.72 & 78.46 & 67.14 & 77.71 & 86.25 & 77.39 \\
        L2S                & 61.54 & 47.94 & 19.58 & 72.92 & 50.49 & 78.27 & 67.14 & 77.08 & 86.46 & 77.24 \\
        Act-Add            & 57.31 & 45.56 & 25.83 & 73.54 & 50.56 & 78.85 & 66.98 & 78.12 & 85.83 & 77.45 \\
        Linear-AcT         & 47.31 & 41.75 & 29.17 & 61.67 & 44.97 & 78.27 & 66.35 & 77.71 & 86.88 & 77.30 \\
        ACT                & 60.38 & 47.78 & 22.29 & 73.75 & 51.05 & 78.46 & 66.83 & 78.12 & 86.46 & 77.47 \\
        RepE               & 56.15 & 47.30 & 25.00 & 75.21 & 50.92 & 78.27 & 67.14 & 78.33 & 86.67 & 77.60 \\
        \hline
        MAST (Ours)        & \underline{79.04} & \underline{53.81} & \underline{39.58} & \textbf{99.58} & \underline{68.00} & \underline{81.15} & \underline{67.62} & \underline{82.71} & \textbf{91.88} & \underline{80.84} \\
        MAST-PR (Ours)     & \textbf{87.69} & \textbf{74.60} & \textbf{51.67} & \underline{90.42} & \textbf{76.09} & \textbf{83.46} & \textbf{68.41} & \textbf{85.83} & \underline{91.46} & \textbf{82.29} \\
        \hline
        \textit{Task-Oracle} & 90.58 & 75.24 & 73.12 & 99.58 & 84.63 & 84.04 & 68.10 & 85.62 & 93.96 & 82.93 \\
        \hline
    \end{tabular}}
    \caption{Accuracy ($\%$) on the four CV-Bench visual grounding tasks (test split), each
    model at the intervention layer chosen by its own layer scan. Best per column
    in bold and second best \underline{underlined}.
    \textit{*}~Norm-Rnd is a norm-matched random-direction control, and
    \textit{Task-Oracle} applies the optimal hard direction weights of the
    ground-truth task as an upper bound in the four-task-average sense, excluded
    from the ranking. Baseline paradigms and frozen hyperparameters are described
    in Sections~\ref{sec:setup} and~\ref{sec:main}. Baseline references are
    CAA~\citep{rimsky-etal-2024-steering},
    L2S~\citep{parekh2025neurips-learning},
    Act-Add~\citep{turner2024steeringlanguagemodelsactivation},
    Linear-AcT~\citep{rodriguez2025controlling},
    ACT~\citep{10.1145/3696410.3714640}, and
    RepE~\citep{zou2025representationengineeringtopdownapproach}.}
    \label{tab:main_cvbench}
\end{table}

MAST achieves a consistent, substantial advantage on visual grounding. On LLaVA-v1.5 it reaches a four-task average of $68.00\%$ ($+17.06$ points over No-steering) and is the only method to surpass every standard baseline on all four tasks. Even tuned to its own best injection strength, no static-direction baseline improves meaningfully over No-steering, a direct manifestation of the cross-cluster conflict of Section~\ref{sec:motivation}. Per-baseline numbers are in Appendix~\ref{sec:appendix_main}.

The advantage transfers to Qwen2-VL. Despite an already high No-steering average of $77.40\%$, MAST attains the best result on all four tasks ($80.84\%$, $+3.44$ points), gaining $5.00$ and $5.21$ points on Distance and Depth. The static baselines are so sensitive to injection strength that per-baseline tuning drives them toward near-zero intervention, whereas MAST regulates its magnitude through adaptive per-cluster shrinkage, adapting to the larger activation scale of Qwen2-VL.

McNemar's paired test against each of the seven baselines confirms genuine per-sample gains, significant on all four LLaVA-v1.5 tasks ($p<0.01$) and on Qwen2-VL Relation, Distance, and Depth, the sole exception being the near-saturated Count. With the zero cross-seed variance of Section~\ref{sec:setup}, the improvements are deterministic. The full test with per-cell discordant-pair counts $b$, $c$, and $p$-values is in Appendix~\ref{sec:appendix_main}.

Task-specialized reweighting further amplifies the gains. MAST-PR raises the LLaVA-v1.5 average from $68.00\%$ to $76.09\%$ ($+8.09$ points), with the largest gain on Count ($53.81\%\!\rightarrow\!74.60\%$), and raises Qwen2-VL from $80.84\%$ to $82.29\%$, near its $82.93\%$ Task-Oracle bound. Thus LLaVA-v1.5 has substantially more room for direction specialization than Qwen2-VL. Section~\ref{sec:ablation} provides controlled subspace comparisons; Appendices~\ref{sec:appendix_main}, \ref{sec:appendix_direction}, and~\ref{sec:appendix_router} provide per-task results, the full direction scan, and robustness controls.

\begin{figure}[htbp]
    \centering
    \begin{minipage}[t]{\FigFivePanelWidth}
        \centering
        \vspace{0pt}
        \includegraphics[width=\linewidth]{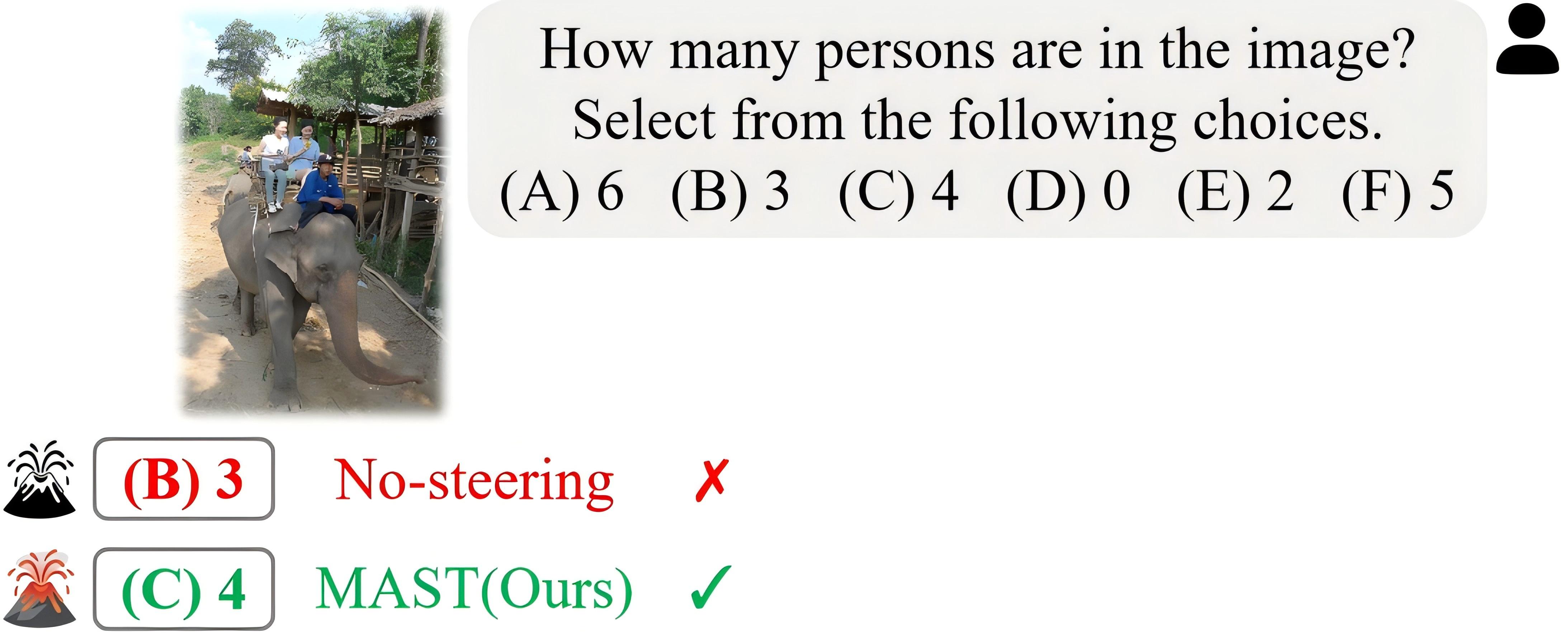}
        \caption{Qualitative Count example. No-steering answers~3, while MAST
        routes the query to its nearest Count cluster and outputs the correct
        answer~4.}
        \label{fig:case_study_count}
    \end{minipage}\hfill
    \begin{minipage}[t]{\PopeTablePanelWidth}
        \centering
        \vspace{0pt}
        \setlength{\tabcolsep}{2pt}
        \renewcommand{\arraystretch}{1.06}
        \resizebox{\linewidth}{!}{%
        \begin{tabular}{lcccccc}
            \hline
            & \multicolumn{2}{c}{Random} & \multicolumn{2}{c}{Popular} & \multicolumn{2}{c}{Adversarial} \\
            \cline{2-3}\cline{4-5}\cline{6-7}
            Method & Acc & F1 & Acc & F1 & Acc & F1 \\
            \hline
            No-steering   & 88.80 & 88.81 & 85.07 & 85.61 & 78.47 & 80.50 \\
            Norm-Rnd      & 88.37 & 88.50 & 84.50 & 85.24 & 77.93 & 80.23 \\
            CAA           & 88.03 & 88.46 & 84.07 & 85.20 & 77.03 & 79.98 \\
            Linear-AcT    & 88.77 & 89.12 & 84.83 & 85.85 & 77.80 & 80.56 \\
            ACT           & \underline{89.33} & \underline{89.40} & \underline{85.60} & \textbf{86.21} & \underline{79.03} & \textbf{81.11} \\
            L2S           & 88.93 & 89.19 & 84.10 & 85.16 & 76.97 & 79.85 \\
            RepE          & 88.47 & 88.63 & 84.73 & 85.48 & 77.97 & 80.31 \\
            \textbf{MAST (Ours)} & \textbf{89.87} & \textbf{89.75} & \textbf{85.80} & \underline{86.20} & \textbf{79.13} & \underline{80.96} \\
            \hline
        \end{tabular}}
        \captionof{table}{Accuracy and F1 ($\%$) on the three POPE test splits
        for LLaVA-v1.5. Best per column in bold and second best
        \underline{underlined}.}
        \label{tab:main_pope}
    \end{minipage}
\end{figure}

Figure~\ref{fig:case_study_count} illustrates one Count instance in which the image contains four persons. No-steering outputs~3, whereas MAST routes the query to a nearby Count prototype and outputs the correct answer,~4. This example is illustrative rather than a causal test of the mechanism; the controlled aggregate evidence is reported in Table~\ref{tab:subspace_controls} and the ablation study. Additional examples are provided in Appendix~\ref{sec:appendix_qualitative}.

Using the shared object-labeled POPE support bank described above, rather than the CV-Bench prototype bank, MAST delivers stable hallucination-suppression gains on POPE (Table~\ref{tab:main_pope}). On LLaVA-v1.5 it surpasses No-steering on the accuracy and F1 of all three splits and attains the best of the six metrics on four, the other two F1 within $0.15$ points of the best baseline ACT. The POPE space is narrower and baselines already high, so gains are smaller than on CV-Bench, but MAST remains robust where some baselines degrade.

\subsection{Ablation Study}
\label{sec:ablation}

To decompose each component's contribution, we remove one at a time on CV-Bench in a leave-one-out manner (Table~\ref{tab:component_ablation}). Multi-center routing drives the main gain of base MAST. We next test whether local bases improve task-specialized reweighting in MAST-PR.

\begin{figure}[htbp]
    \centering
    \begin{minipage}[t]{\AblationTablePanelWidth}
        \centering
        \vspace{0pt}
        \resizebox{\linewidth}{!}{%
        \begin{tabular}{lccccr}
            \hline
            Variant & Md & Mc & LB & Sh & Average \\
            \hline
            \textbf{MAST-PR} (Ours) & \checkmark & \checkmark & \checkmark & \checkmark & \textbf{76.09} \\
            \hline
            \textbf{MAST} (Ours) & \checkmark & \checkmark & \checkmark & \checkmark & \textbf{68.00} \\
            \;w/o Shrinkage      & \checkmark & \checkmark & \checkmark & $\times$ & 67.00 \\
            \;w/o Local Basis    & \checkmark & \checkmark & $\times$ & \checkmark & 67.67 \\
            \;w/o Multi-center   & \checkmark & $\times$ & \checkmark & \checkmark & 39.54 \\
            \;w/o Medoid         & $\times$ & $\times$ & $\times$ & $\times$ & 47.35 \\
            \hline
        \end{tabular}}
        \captionof{table}{Leave-one-out component ablation on CV-Bench (test
        split), four-task average. Per-task columns in
        Appendix~\ref{sec:appendix_ablation}. \textit{*}~Md is medoid
        prototype, Mc multi-center routing, LB local basis, and Sh adaptive
        shrinkage. The MAST block removes one component at a time under equal
        weighting.}
        \label{tab:component_ablation}
    \end{minipage}\hfill
    \begin{minipage}[t]{\SubspaceTablePanelWidth}
        \centering
        \vspace{0pt}
        \setlength{\tabcolsep}{2pt}
        \renewcommand{\arraystretch}{1.06}
        \resizebox{\linewidth}{!}{%
        \begin{tabular}{lcccr}
            \hline
            Representation & Local & \shortstack{Data-\\driven} & \shortstack{Task-\\cond.} & Average \\
            \hline
            Equal-weight & \checkmark & \checkmark & $\times$ & 68.00 \\
            Random basis & \checkmark & $\times$ & \checkmark & 68.20 \\
            Raw directions & \checkmark & \checkmark & \checkmark & 74.97 \\
            Global PCA & $\times$ & \checkmark & \checkmark & 77.67 \\
            \textbf{Local PCA/SVD} & \checkmark & \checkmark & \checkmark & \textbf{84.05} \\
            \hline
        \end{tabular}}
        \captionof{table}{Controlled task-oracle comparison on LLaVA-v1.5
        CV-Bench. All rows share the MAST backbone, medoid routing, $K{=}5$,
        $r{=}8$, layer, support protocol, and evaluation. For task-conditioned
        rows, ground-truth task labels are used only to select a subspace
        direction and weight on a disjoint validation split, which are frozen
        before testing.}
        \label{tab:subspace_controls}
    \end{minipage}
\end{figure}

Under fixed MAST backbone, medoid routing, local rank, layer, support protocol,
and evaluation, local PCA/SVD subspaces reach $84.05\%$, outperforming global
PCA ($77.67\%$), raw local directions ($74.97\%$), and random local bases
($68.20\%$). This result supports the use of local subspace directions to form
task-specialized direction families in MAST-PR.

The $39.54\%$ of w/o~Multi-center falls even below the $47.35\%$ of w/o~Medoid, which strips every component and degenerates to the global mean. Once collapsed to a single cluster, the local basis and shrinkage lose the local geometry they model, so both are refinement mechanisms coupled to multi-center routing rather than independent components. The degeneration mechanism is detailed in Appendix~\ref{sec:appendix_ablation}.

\subsection{Effect of Intervention Layer}
\label{sec:layer}

\begin{figure}[htbp]
    \centering
    \includegraphics[width=0.95\linewidth]{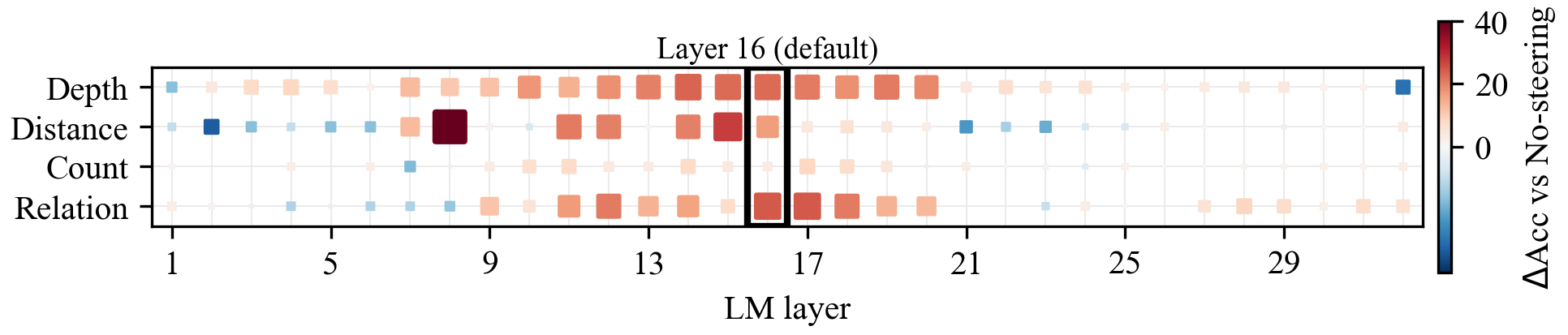}
    \caption{CV-Bench accuracy gain of MAST over No-steering at every
    language-model (LM) layer of LLaVA-v1.5 (validation split). Layer~16
    (black box) is the default intervention layer.}
    \label{fig:layer_scan_heatmap}
\end{figure}

MAST's gains concentrate in the mid-to-late LM. To keep the layer choice free of test-set information, we scan all $32$ LM layers of LLaVA-v1.5 and freeze the selected layer for testing. Positive four-task gain spans layers~11 to~20, within which layers~12, 14, 15, 16 form a plateau varying by less than $1.60$ points (Figure~\ref{fig:layer_scan_heatmap}). We fix layer~16 (within $0.03$ points of the validation maximum), so the default needs no fine tuning within this plateau.

The per-task optima are close but not identical. Under the unified layer, the geometric tasks sit slightly off their own optima, leaving room for the direction-level specialization exploited by MAST-PR. Full per-task analysis, including a sharp Distance-only spike at layer~8, is in Appendix~\ref{sec:appendix_layer}.

\subsection{Hyperparameter Sensitivity}
\label{sec:hyperparam}

One-at-a-time validation sweeps expose two trade-offs. Increasing the cluster count eventually fragments Distance support, motivating $K{=}5$; $k{=}1$ maximizes the average ($72.26\%$) but collapses Distance to $26.67\%$, motivating $k{=}3$. Performance is comparatively insensitive to $\alpha_{\max}$ and rank $r$. Direction-family optima also differ across tasks and models. On LLaVA-v1.5, Distance, Count, and Relation prefer ranks 2, 5, and 6, while Depth prefers equal weighting. Overweighting a non-preferred direction family can sharply collapse accuracy, motivating task-specific frozen weights. Complete curves and the weight$\times$direction tables are in Appendices~\ref{sec:appendix_hyperparam} and~\ref{sec:appendix_direction}.

\paragraph{Efficiency.} Under the default $K{=}5$, $r{=}8$, and $k{=}3$ configuration, MAST requires a one-time $3.93$-second incremental bank construction and stores a $0.938$-MiB fp32 bank, versus a $16$-KiB global steering vector; MAST-PR's retained local bases increase this to $3.438$~MiB. Local SVD is offline-only, and GPU-synchronized query-to-medoid routing plus top-$k$ aggregation costs $0.65$~$\mu$s per query for base MAST. The complete protocol and prototype-count scaling are in Appendix~\ref{sec:appendix_efficiency}.

\section{Conclusion}

We recast activation steering in MLLMs through activation geometry, where visual-task activations form separated clusters, the intra-cluster signal is low-rank, and cross-cluster directions nearly cancel under averaging, so a single global-mean direction fails on heterogeneous tasks. MAST anchors a medoid to each cluster, routes each query to its nearest few, extracts a local low-rank direction, and aggregates adaptively, while MAST-PR reweights task-specialized local-direction families per query task. On CV-Bench and POPE, MAST improves the four-task average by $17.06$ points over No-steering and MAST-PR adds $8.09$ points. Controlled task-oracle comparisons further show that local subspace directions support task-specialized steering more effectively than global, raw, or random alternatives.

\subsection*{AI use statement}
Generative AI tools were used only for research ideation, drafting and polishing
the manuscript, ICLR format adaptation, and typesetting checks. They were not
used for any required research activity, including methodology or experiment
design, method implementation, data processing, mathematical claims or proofs,
or result interpretation. All AI-assisted material was reviewed by the authors,
who take full responsibility for the final manuscript.


\bibliography{references}
\bibliographystyle{iclr2027_conference}

\appendix
\setcounter{figure}{0}
\setcounter{table}{0}
\renewcommand{\thefigure}{S\arabic{figure}}
\renewcommand{\thetable}{S\arabic{table}}
\renewcommand{\theHfigure}{S\arabic{figure}}
\renewcommand{\theHtable}{S\arabic{table}}
\input{appendix}

\end{document}

%% file: appendix.tex
\section{Heterogeneous Geometry}
\label{sec:appendix_geometry}

This section provides the complete statistics and the full four-task
visualizations behind the three observations of the main paper's section on
heterogeneous activation geometry. All statistics are computed at layer~16 of
LLaVA-v1.5 with label-grouped $K$-medoids clustering ($K{=}5$ per task). For
each observation we restate the claim and report the supporting numbers the
main paper defers here.

\begin{figure}[htbp]
    \centering
    \includegraphics[width=0.60\linewidth]{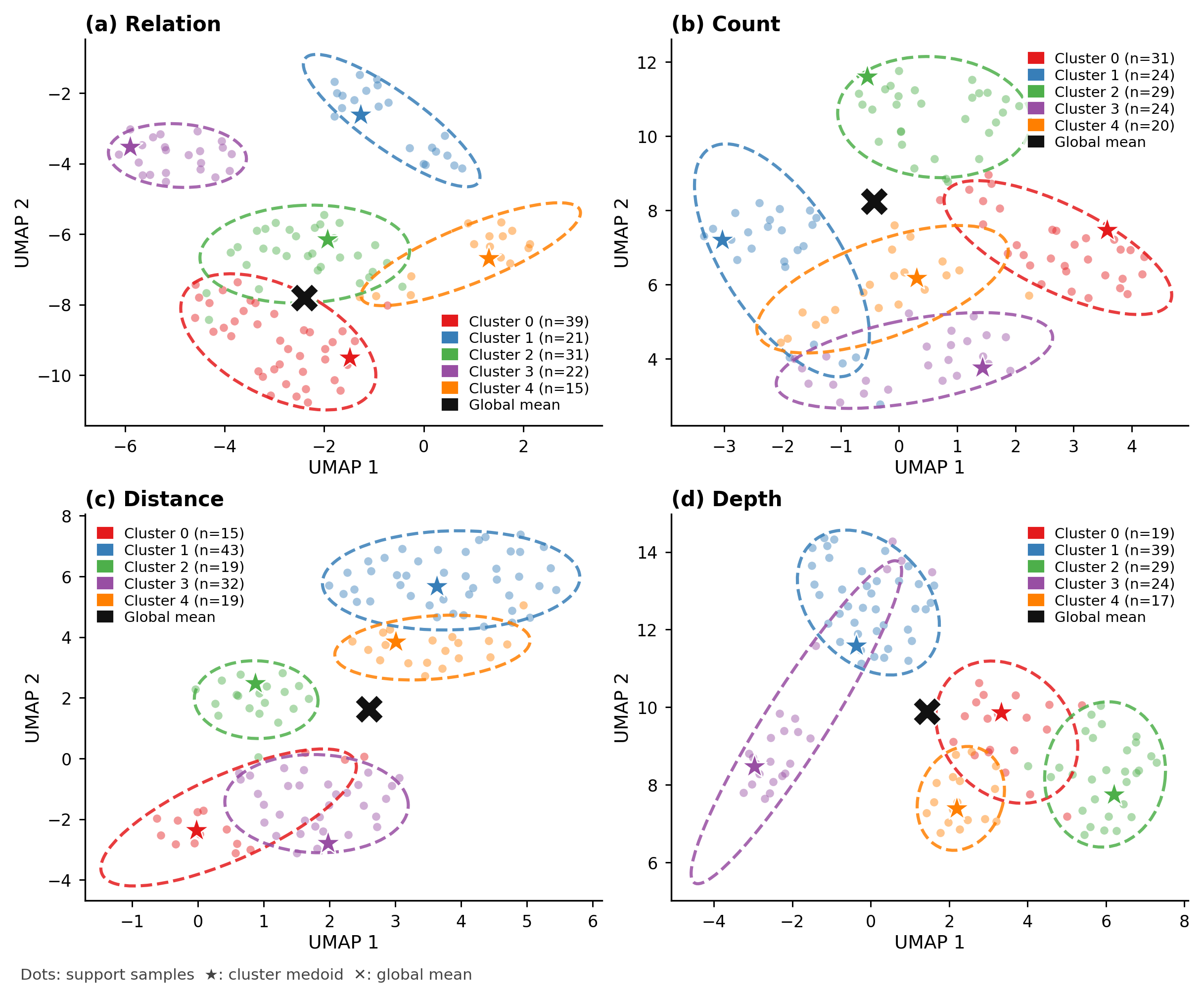}
    \caption{Full four-task UMAP projection of CV-Bench support activations.
    Support samples are colored by cluster, stars mark the cluster medoids, and
    the black cross marks the global mean. Across all four tasks the global mean
    lands in the empty region between clusters, confirming that the multi-center
    geometry is a general phenomenon.}
    \label{fig:obs1_pca_scatter_all}
\end{figure}

\textbf{Observation~1: activations form multiple local clusters.}
Table~\ref{tab:clustering_quality} reports the full clustering-quality metrics.
The per-task cluster separability ranges from $0.144$ (Count) to $0.192$
(Relation) and the Davies-Bouldin index from $1.769$ (Relation) to $2.142$
(Distance). Across all four tasks these values indicate that the clusters are
mutually separated rather than overlapping. Per-task nearest-medoid routing is
perfect ($100\%$ top-1), and even under the combined 20-cluster setting over the
four tasks the routing accuracy stays at $76.69\%$, far above the $20\%$ random
baseline, confirming that the spatial layout of the clusters already encodes
task identity. Figure~\ref{fig:obs1_pca_scatter_all} shows the full four-task
UMAP projection: in every task the global mean (black cross) lands in the empty
region between clusters, so the multi-center geometry is general rather than
specific to the Distance panel shown in the main paper.

\begin{table}[htbp]
    \centering
    {\small
    \setlength{\tabcolsep}{3pt}
    \begin{tabular}{llrrrr}
        \hline
        Scope & Task & Clusters & Separability $\uparrow$ & DB $\downarrow$ & Top-1 \\
        \hline
        Per-task & Relation      &  5 & 0.192 & 1.769 & 100.00 \\
        Per-task & Count         &  5 & 0.144 & 2.051 & 100.00 \\
        Per-task & Distance      &  5 & 0.159 & 2.142 & 100.00 \\
        Per-task & Depth         &  5 & 0.170 & 1.978 & 100.00 \\
        All      & all (4 tasks) & 20 & 0.150 & 2.035 &  76.69 \\
        \hline
    \end{tabular}}
    \caption{Clustering quality of the label-grouped $K$-medoids clusters.
    Separability (higher is better), DB is the Davies-Bouldin index
    (lower is better), and Top-1 is the top-1 routed same-task accuracy. Per-task
    routing is perfect, and the combined 20-cluster setting stays far above the
    $20\%$ random baseline.}
    \label{tab:clustering_quality}
\end{table}

\begin{figure}[htbp]
    \centering
    \includegraphics[width=0.60\linewidth]{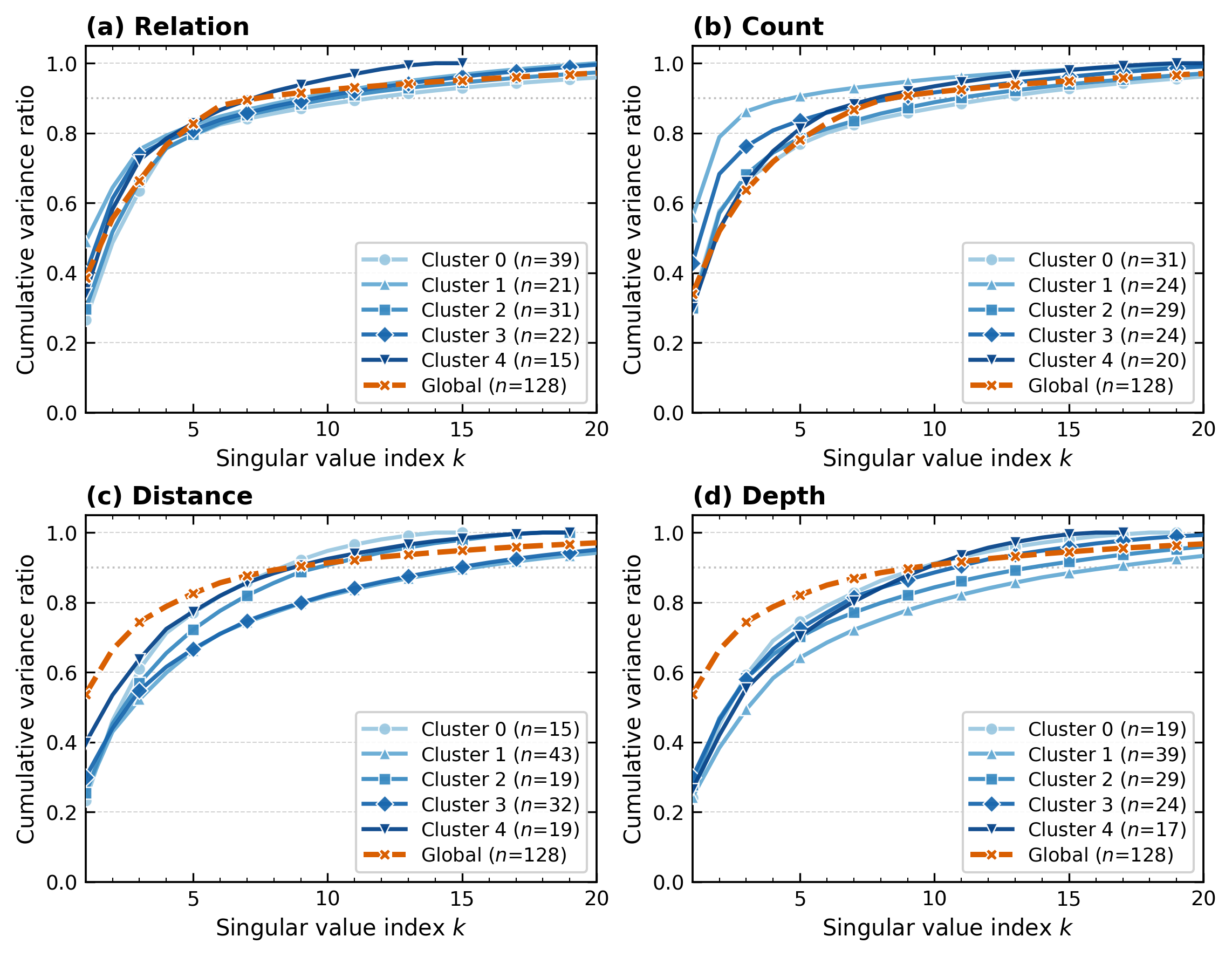}
    \caption{Full four-task cumulative variance ratio of the singular value
    decomposition of the intra-cluster steering vectors. Solid lines denote
    per-cluster curves, the black dashed line the global baseline without
    clustering, and the dotted grey line the $90\%$ threshold. Across all tasks
    the leading singular directions capture at least $90\%$ of the variance well
    below the ambient dimension, confirming that the intra-cluster low-rank
    structure is a general property.}
    \label{fig:singular_value_cumvar_all}
\end{figure}

\textbf{Observation~2: the intra-cluster steering signal is low-rank.}
Figure~\ref{fig:singular_value_cumvar_all} gives the full four-task
cumulative-variance curves of the singular value decomposition of the
intra-cluster steering vectors. Across all clusters of all four tasks the leading
$17$ singular directions already capture no less than $90\%$ of the variance,
whereas the ambient activation dimension is $4096$. The most concentrated Count
cluster needs only five directions to reach the same threshold. The black dashed
baseline, computed without clustering, sits far below the per-cluster curves
because its leading direction is merely the average of the conflicting
per-cluster directions and is nearly orthogonal to each of them. This low rank is
what lets MAST truncate a clean local direction almost without loss, and it is
corroborated by the rank sweep in Figure~\ref{fig:alpha_r_sensitivity} and
Table~\ref{tab:hyperparam_sensitivity} (Section~\ref{sec:appendix_hyperparam}),
where the four-task average is nearly flat in the local basis rank $r$: sweeping
$r$ from $1$ to $4096$ changes the average by only $0.76$ points, so retaining
only the first few singular directions already approaches the full-rank average.

\begin{figure}[htbp]
    \centering
    \includegraphics[width=0.60\linewidth]{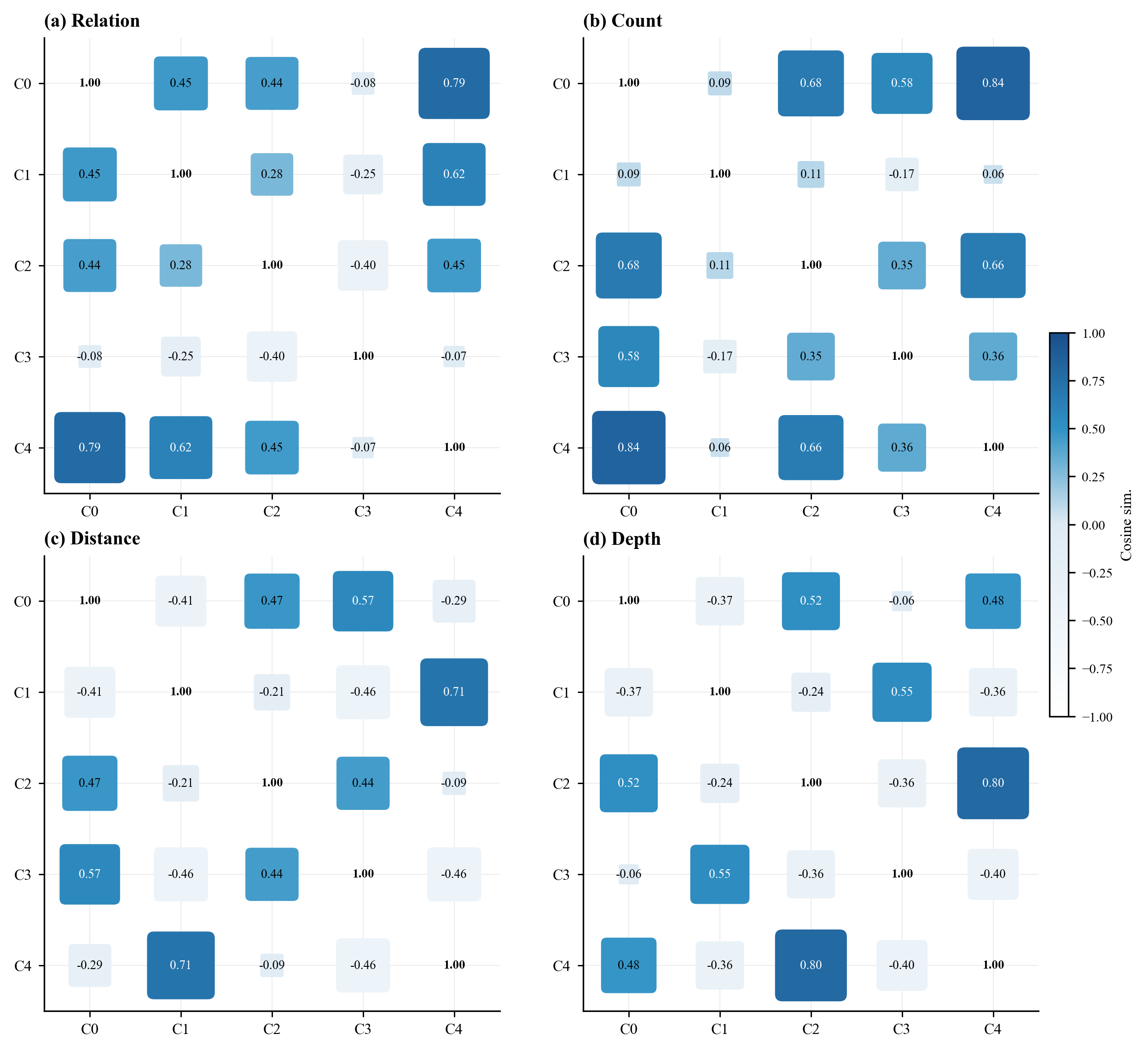}
    \caption{Full four-task pairwise cosine similarity between the five cluster
    medoid steering directions within each CV-Bench task. Red denotes negative
    correlation and blue positive. The geometric tasks Distance and Depth exhibit
    the most severe near-orthogonal conflict among cluster directions, while
    Relation and Count show a less extreme but still substantial directional
    spread.}
    \label{fig:obs3_cosine_heatmap_2x2}
\end{figure}

\begin{figure}[htbp]
    \centering
    \includegraphics[width=0.60\linewidth]{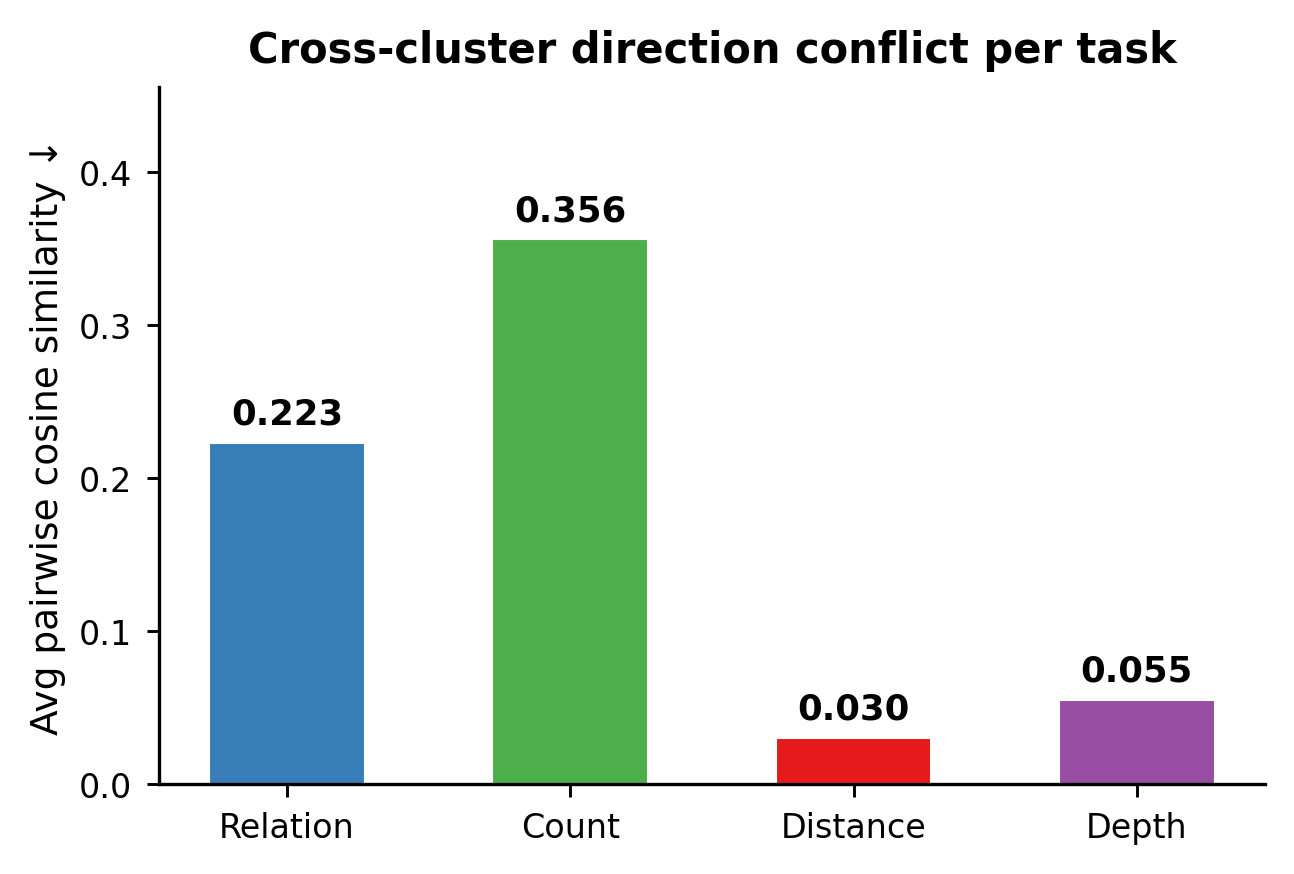}
    \caption{Average intra-task pairwise cosine similarity between cluster
    steering directions across the four CV-Bench tasks. Lower means more severe
    directional conflict, lowest for Depth and Distance.}
    \label{fig:obs3_avg_pairwise_bar}
\end{figure}

\textbf{Observation~3: cross-cluster directions conflict.}
Table~\ref{tab:cosine_similarity} reports, for each task, the cosine similarity
between every cluster's steering direction and the global mean direction. The
per-cluster averages are $0.576$ (Relation), $0.654$ (Count), $0.418$
(Distance), and $0.364$ (Depth), so the global mean never exceeds a similarity of
$0.66$ with the clusters it is meant to summarize and is faithful to none of
them. The individual entries expose the outliers that pull the mean off course:
Relation contains a cluster negatively correlated with the global mean (C3 at
$-0.106$), and Count contains an outlier cluster whose similarity is only
$0.063$. Figure~\ref{fig:obs3_cosine_heatmap_2x2} shows the full four-task
\emph{pairwise} similarity between the five cluster medoid directions: the
geometric tasks Distance and Depth are almost mutually orthogonal (average
pairwise similarity $0.030$ and $0.055$), the Relation task contains an opposing
cluster whose pairwise similarities range from $-0.08$ to $-0.40$, and even the
most coherent task Count has an outlier cluster below $0.12$ with every other
cluster. This near-orthogonality is why the linearly averaged global mean shrinks
in magnitude and loses steering capacity, and it is exactly the geometry that
MAST avoids by never averaging across conflicting clusters.
Figure~\ref{fig:obs3_avg_pairwise_bar} summarizes the average intra-task
pairwise similarity for all four tasks, ordering them by severity of directional
conflict: it is lowest for the geometric tasks Depth and Distance and highest for
Count, matching the per-cluster and pairwise values above.

\begin{table}[htbp]
    \centering
    {\small
    \setlength{\tabcolsep}{4pt}
    \begin{tabular}{lrrrrrr}
        \hline
        Task & C0 & C1 & C2 & C3 & C4 & Average \\
        \hline
        Relation & 0.839 & 0.750 & 0.573 & -0.106 & 0.823 & 0.576 \\
        Count    & 0.895 & 0.063 & 0.812 &  0.696 & 0.802 & 0.654 \\
        Distance & 0.387 & 0.337 & 0.491 &  0.407 & 0.466 & 0.418 \\
        Depth    & 0.518 & 0.243 & 0.364 &  0.439 & 0.254 & 0.364 \\
        \hline
    \end{tabular}}
    \caption{Cosine similarity between each cluster's steering direction and the
    global mean direction, per CV-Bench task. The final column averages over the
    five clusters. The global mean stays far from every local cluster direction,
    confirming that it faithfully represents none of them.}
    \label{tab:cosine_similarity}
\end{table}

\section{Method Details}
\label{sec:appendix_method}

For offline direction extraction, two teacher-forced passes of $P_i$ with
$y_i^{+}$ and $y_i^{-}$ give the mean completion states $a_i^{+}$ and
$a_i^{-}$, hence $\delta_i=a_i^{+}-a_i^{-}$. Routing instead reads the
answer-free last-prompt state $c_i$. For an online query $P_q$, the corresponding
state $c_q$ is scored by $s_{q,c}=\cos(c_q,c_{m_c})$ to construct $v$. A
layer-$\ell$ residual-stream hook adds $v$ to all merged prompt positions in the
initial forward pass and to every subsequently generated token; the query never
contains a gold or candidate answer.

This section gives the reliability and blending formulas omitted from the local
low-rank steering and adaptive aggregation of the main paper. For each selected
cluster we build its local reliability $\rho_c$ from three signals in $[0,1]$,
relative cluster size, compactness, and routing confidence, where
$R_c=\frac{1}{|c|}\sum_{i\in c}\lVert c_i-c_{m_c}\rVert_2$ is the cluster
radius ($0$ for a singleton), $\bar R$ its mean over all prototypes, and
$s_{q,c}=\cos(c_q,c_{m_c})$ the raw routing similarity before temperature
scaling,
\begin{equation}
\begin{aligned}
\rho_c^{\text{size}}&=\frac{|c|}{\max_{c'}|c'|}\\
\rho_c^{\text{cmp}}&=\frac{1}{1+R_c/\bar R}\\
\rho_c^{\text{conf}}&=\frac{\mathrm{clip}(s_{q,c},-1,1)+1}{2}
\end{aligned}
\end{equation}
A fixed convex combination with weights
$(\lambda_{\text{size}},\lambda_{\text{cmp}},\lambda_{\text{conf}})=(0.4,0.3,0.3)$
gives
\begin{equation}
\rho_c=\frac{\lambda_{\text{size}}\rho_c^{\text{size}}+\lambda_{\text{cmp}}\rho_c^{\text{cmp}}+\lambda_{\text{conf}}\rho_c^{\text{conf}}}{\lambda_{\text{size}}+\lambda_{\text{cmp}}+\lambda_{\text{conf}}}
\end{equation}
The blending coefficient $\alpha_c$ decreases with reliability, clipped to
$[\alpha_{\min},\alpha_{\max}]=[0.05,0.80]$,
\begin{equation}
\alpha_c=\alpha_{\max}-(\alpha_{\max}-\alpha_{\min})\,\rho_c
\end{equation}
yielding the base-MAST intra-cluster blend
$v_c=(1-\alpha_c)\,\delta_{m_c}+\alpha_c\,\delta_c^{\mathrm{MAST}}$ of the main
paper. When reliability is low, $\alpha_c$ is large and the direction leans on
the denoised reconstruction, suppressing noise from inconsistent intra-cluster
directions. When reliability is high, more of the medoid's raw signal is kept.

\subsection{Complete MAST Procedure}

Algorithm~\ref{alg:mast} collects the offline bank construction and online
routing steps that are described algebraically in the main paper.

\begin{algorithm}[htbp]
\caption{MAST}
\label{alg:mast}
\begin{algorithmic}[1]
\REQUIRE support set $\{\mathcal{S}_\tau\}$, answer-free query context $c_q$, hyperparameters $K,r,k,T,\alpha_{\min},\alpha_{\max}$
\STATE \textbf{Offline.} For each task run $K$-medoids on $\{c_i\}$. For each cluster $c$ store the medoid $(c_{m_c},\delta_{m_c})$, local basis $B_c$ (top-$r$ SVD of centered residuals), $\bar\delta_c$, $\delta_c^{\mathrm{MAST}}=\tilde\delta_c(\mathbf{1}_r)$, and cluster size and compactness
\STATE \textbf{Online.} $s_{q,c}=\cos(c_q,c_{m_c})$. Take top-$k$ as $\mathcal{N}_k(q)$. Set $w_c=\mathrm{softmax}(s_{q,c}/T)$
\FOR{$c \in \mathcal{N}_k(q)$}
    \STATE $\rho_c=\mathrm{reliability}(\text{size, compactness, routing conf.})$. Set $\alpha_c=\alpha_{\max}-(\alpha_{\max}-\alpha_{\min})\rho_c$ and $v_c=(1-\alpha_c)\delta_{m_c}+\alpha_c\delta_c^{\mathrm{MAST}}$
\ENDFOR
\ENSURE steering vector $v=\sum_{c\in\mathcal{N}_k(q)} w_c\,v_c$, injected by $a\leftarrow a+v$
\end{algorithmic}
\end{algorithm}

\subsection{Complete MAST-PR Reweighting Rule}
\label{sec:appendix_mastpr}

For rank $j$, let $\mathcal{F}_j=\{b_{c,j}:c\in\mathcal{C}\}$ denote the
family of $j$-th local-basis directions across the prototype bank. We call
$\mathcal{F}_j$ \emph{task-specialized} for task $\tau$ when emphasizing its
members yields the strongest validation performance for $\tau$. This is an
operational definition: the independently computed bases remain cluster-local
coordinates, and it does not imply that a task-pure cluster contains multiple
task identities.

A four-way logistic-regression probe maps $c_q$ to $p(\tau\mid c_q)$. The
direction-family rank $j_\tau$ and scalar weight $w_\tau$ are selected offline
on the validation split. With $\sigma_{c,j}$ the $j$-th local singular value,
the selected per-cluster weight is
\begin{equation}
\beta_{c,j}^{\tau}=w_\tau\frac{\sigma_{c,1}}
{\max(\sigma_{c,j},10^{-12})}\mathbf{1}[j=j_\tau].
\end{equation}
All remaining directions are zeroed. For Depth, where no single family beats
equal weighting, $\beta_{c,j}^{\mathrm{Depth}}=1$ for every retained $j$. At
inference, MAST-PR softly mixes these weights,
$\beta_{q,c,j}=\sum_\tau p(\tau\mid c_q)\beta_{c,j}^{\tau}$, reconstructs
$\delta_{q,c}^{\mathrm{PR}}$ as in the main paper, and substitutes it for
$\delta_c^{\mathrm{MAST}}$ in the unchanged blend and aggregation pipeline.

\section{Task-Probe Reliability and Robustness of MAST-PR}
\label{sec:appendix_router}

This section studies only the auxiliary MAST-PR task probe used to mix
task-specialized direction weights. The prototype bank, nearest-medoid routing,
local bases, and all MAST hyperparameters remain fixed throughout; thus these
interventions do not alter MAST's multi-center medoid routing.

\begin{table}[htbp]
    \centering
    {\small
    \setlength{\tabcolsep}{5pt}
    \begin{tabular}{lccccc}
        \hline
        True\,$\backslash$\,Pred. & Relation & Count & Distance & Depth & Recall \\
        \hline
        Relation & \textbf{63.54} & 5.09  & 20.16 & 11.20 & 100.00 \\
        Count    & 4.14  & \textbf{92.84} & 1.61  & 1.41  & 100.00 \\
        Distance & 27.26 & 6.05  & \textbf{58.88} & 7.81  & 100.00 \\
        Depth    & 11.62 & 2.26  & \underline{49.62} & \textbf{36.50} & 18.67 \\
        \hline
    \end{tabular}}
    \caption{Soft routing-probability matrix of the MAST-PR task probe on the
    CV-Bench test split (LLaVA-v1.5, layer~16). Row $i$, column $j$ is the mean
    probability assigned to task $j$ over queries whose true task is $i$, diagonal
    entries (bold) the mass kept on the correct task, and the rightmost
    column auxiliary top-1 task-classification accuracy.}
    \label{tab:router_confusion}
\end{table}

The probe, trained on the support set ($99.80\%$ training accuracy), assigns high
mean probability to Relation, Count, and Distance. On Depth, it assigns only
$36.50\%$ mean probability to the correct task and $49.62\%$ to Distance. Since
MAST-PR uses a probability-weighted mixture of task-specialized direction
families, this uncertainty places substantial weight on the Distance-specialized
family and is
associated with Depth's drop from $99.58\%$ to $90.42\%$ in the main-paper
CV-Bench results. The $18.67\%$ top-1 accuracy is an auxiliary diagnostic, not
the inference routing rule.

\paragraph{Uniform posterior-noise robustness.}
We perturb only the test-time task posterior as
$p_{\epsilon}(\tau\mid c_q)=(1-\epsilon)p(\tau\mid c_q)+\epsilon/4$, without
retraining the probe or changing any steering asset. Table~\ref{tab:probe_noise}
shows that the reweighting gain remains positive through $\epsilon=0.75$ and
falls slightly below equal-weight MAST only when the posterior is completely
uniform. The small change at $\epsilon=0.25$ is not interpreted as a benefit of
noise. A uniform mixture of task-specialized weights is not identical to
equal-weight MAST, explaining why $\epsilon=1$ need not equal the MAST row.

\begin{table}[htbp]
    \centering
    \small
    \setlength{\tabcolsep}{6pt}
    \begin{tabular}{lrrrrr}
        \hline
        Posterior noise $\epsilon$ & Relation & Count & Distance & Depth & Average \\
        \hline
        0.00 & 87.69 & 74.60 & 51.67 & 90.42 & 76.09 \\
        0.25 & 92.69 & 68.41 & 47.29 & 96.88 & \textbf{76.32} \\
        0.50 & 95.00 & 59.05 & 37.08 & 97.92 & 72.26 \\
        0.75 & 94.23 & 52.38 & 35.21 & 98.75 & 70.14 \\
        1.00 & 91.54 & 47.94 & 30.63 & 98.12 & 67.06 \\
        \hline
        Equal-weight MAST & 79.04 & 53.81 & 39.58 & 99.58 & 68.00 \\
        \hline
    \end{tabular}
    \caption{Uniform task-posterior perturbation of MAST-PR on the LLaVA-v1.5
    CV-Bench test split. All rows retain identical medoid routing and steering
    assets; only the task-probe posterior used for direction reweighting changes.}
    \label{tab:probe_noise}
\end{table}

\paragraph{Counterfactual Depth-to-Distance intervention.}
To test the consequence of the observed leakage direction, we evaluate the
Depth test queries after replacing their task-specific weights with the
Distance-specialized weights. This is a counterfactual worst-case intervention,
not a simulation of the empirical $49.62\%$ mean leakage rate. Table~\ref{tab:depth_distance}
shows that the exact wrong specialization is strongly harmful, isolating the
failure mode from medoid routing.

\begin{table}[htbp]
    \centering
    \small
    \setlength{\tabcolsep}{5pt}
    \begin{tabular}{lr}
        \hline
        Depth weighting rule & Accuracy \\
        \hline
        Equal-weight MAST & 99.58 \\
        Native soft MAST-PR & 90.42 \\
        Force Distance-specialized weights & 12.08 \\
        \hline
    \end{tabular}
    \caption{Counterfactual intervention on Depth (LLaVA-v1.5). Relative to
    native soft MAST-PR, forcing Distance-specialized weights lowers accuracy by
    $78.34$ points while leaving medoid routing unchanged.}
    \label{tab:depth_distance}
\end{table}

\paragraph{Fixed confidence-gated safety policy.}
For a conservative deployment policy, we fix $t=0.60$ \emph{a priori}, without
test-set tuning, and interpolate the soft task-specialized weights with
equal-weight MAST using
\begin{equation}
\begin{aligned}
\tilde\beta_{q,c,j}&=\lambda(q)\beta_{q,c,j}
  +[1-\lambda(q)]\mathbf{1}[j\le r_c],\\
\lambda(q)&=\left[\frac{\max_{\tau}p(\tau\mid c_q)-t}{1-t}\right]_0^1.
\end{aligned}
\end{equation}
Table~\ref{tab:probe_gate} shows that this fixed policy recovers Depth to the
equal-weight MAST level while retaining a $5.83$-point average gain over MAST.

\begin{table}[htbp]
    \centering
    \small
    \setlength{\tabcolsep}{6pt}
    \begin{tabular}{lrrrrr}
        \hline
        Method & Relation & Count & Distance & Depth & Average \\
        \hline
        Equal-weight MAST & 79.04 & 53.81 & 39.58 & 99.58 & 68.00 \\
        Native soft MAST-PR & \textbf{87.69} & \textbf{74.60} & \textbf{51.67} & 90.42 & \textbf{76.09} \\
        MAST-PR-Safe (fixed $t=0.60$) & 84.23 & 72.54 & 38.96 & \textbf{99.58} & 73.83 \\
        \hline
    \end{tabular}
    \caption{Fixed confidence-gated MAST-PR safety policy on LLaVA-v1.5
    CV-Bench. The threshold is a pre-specified conservative policy, not a
    test-set-selected hyperparameter.}
    \label{tab:probe_gate}
\end{table}

\section{Full Experimental Setup}
\label{sec:appendix_setup}

This section provides the dataset sizes, the complete hyperparameter
configuration, the baseline tuning protocol, and the implementation and
determinism details deferred from the experimental setup of the main paper.

\textbf{Datasets and metrics.} The official CV-Bench test set contains $2{,}638$
samples, which our task-stratified partition divides into a held-out validation split
of $528$ samples ($20\%$), used for every model-selection decision, and a
disjoint test split of $2{,}110$ samples ($80\%$), on which we report per-task and
four-task-average accuracy. POPE contains $9{,}000$ samples over its three splits
(Random, Popular, Adversarial), on which we report accuracy and F1. All methods
share a support set drawn from the COCO training set, $128$ samples per task for
CV-Bench and $1{,}200$ POPE samples ($400$ from each training subset), with no
sample overlap with the test split. For POPE, we pool all $1{,}200$ training
samples into one support bank, group them by target object, and run $K{=}5$
medoids within each object group; the resulting bank is shared by all three test
splits. POPE support images use COCO train2014 and test images use COCO val2014;
no CV-Bench support example or task label enters the POPE bank.

\textbf{Hyperparameters.} MAST's default hyperparameters are $K{=}5$, $k{=}3$,
$r{=}8$, routing temperature $T{=}0.5$, shrinkage bounds $\alpha_{\min}{=}0.05$
and $\alpha_{\max}{=}0.8$, and reliability weights
$(\lambda_{\text{size}},\lambda_{\text{cmp}},\lambda_{\text{conf}})=(0.4,0.3,0.3)$.
All tunable choices, the intervention layer, $K$, $k$, $r$, $\alpha_{\max}$, and
the MAST-PR direction weights, are selected solely on the CV-Bench validation
split and frozen before the test split is touched, and MAST is never tuned per
method.

\textbf{Baseline tuning.} To make the baseline comparison as strong as possible
rather than forcing a shared $\alpha{=}1$, each CV-Bench steering baseline (CAA,
Act-Add, Linear-AcT, ACT) has its injection strength tuned per baseline and per
model on the CV-Bench validation split over $\alpha\in\{0.2,0.5,1,2,5\}$ and is
reported at its best-performing $\alpha$. L2S is a learned steering network and
is kept at its native configuration.

\textbf{Implementation and determinism.} Experiments run on a single node with
eight NVIDIA GeForce RTX~4090 GPUs ($24$~GB each, driver~550.135), dual Intel
Xeon Platinum 8458P CPUs, and $503$~GB of RAM, under Red Hat Enterprise
Linux~7.9 with Python~3.9, PyTorch~2.6.0 (CUDA~12.4), Hugging Face
Transformers~4.57.6, scikit-learn~1.6.1, and NumPy~2.0.2, in fp16 with greedy
decoding and a maximum generation length of~16. Because
decoding is greedy and the support set is fixed, every method is deterministic
given its support set, and MAST's only stochastic element is the $K$-medoids
initialization. Re-running the full CV-Bench evaluation under three different
clustering seeds ($0$, $42$, $123$) leaves every per-task accuracy unchanged,
with a standard deviation of $0.00$ across seeds on both models
(Table~\ref{tab:multiseed}). We therefore report single-run accuracies and
separately assess statistical reliability with McNemar's paired test against
every baseline on the shared test samples, reported in the main paper.

\begin{table}[htbp]
    \centering
    \setlength{\tabcolsep}{5pt}
    \begin{tabular}{lrrrr}
        \hline
        Model & Relation & Count & Distance & Depth \\
        \hline
        LLaVA-v1.5    & $79.04$ & $53.81$ & $39.58$ & $99.58$ \\
        Qwen2-VL      & $81.15$ & $67.62$ & $82.71$ & $91.88$ \\
        \hline
    \end{tabular}
    \caption{Cross-seed stability of MAST on CV-Bench (test split). Each entry
    is the mean per-task accuracy ($\%$) of MAST over three $K$-medoids
    clustering seeds ($0$, $42$, $123$). The standard deviation across seeds is
    $0.00$ in every cell on both models, so the means coincide exactly with the
    single-run MAST accuracies in the main-paper CV-Bench table.}
    \label{tab:multiseed}
\end{table}

\section{Efficiency and Scaling}
\label{sec:appendix_efficiency}

We quantify the incremental cost of MAST relative to global mean steering on
LLaVA-v1.5 at layer~16. Contrastive support-feature extraction is excluded:
it is required by both methods to construct their steering directions. We time
only MAST's additional bank construction (task-wise clustering and local SVD),
and benchmark its online vector kernel separately from MLLM generation. The
latter uses GPU synchronization and includes cosine routing over all prototypes,
top-$k$ selection, and direction aggregation; it therefore should not be
interpreted as end-to-end generation latency.

\begin{table}[htbp]
    \centering
    \small
    \setlength{\tabcolsep}{4pt}
    \begin{tabular}{lrrrrr}
        \hline
        Method & $K$ & $C$ & Offline (s) & Online ($\mu$s/q.) & Bank (MiB) \\
        \hline
        Global mean & -- & 1 & -- & -- & 0.016 \\
        MAST & 1  & 4  & 0.281 & 0.72 & 0.188 \\
        MAST & 5  & 20 & 3.930 & 0.65 & 0.938 \\
        MAST & 10 & 40 & 4.071 & 0.68 & 1.875 \\
        MAST & 20 & 80 & 4.703 & 0.66 & 3.750 \\
        MAST-PR & 5 & 20 & shared & -- & 3.438 \\
        \hline
    \end{tabular}
    \caption{Incremental efficiency cost on LLaVA-v1.5 CV-Bench support and
    test contexts in fp32. $K$ is medoids per task and $C{=}4K$ the total
    prototype count. ``Online'' is the base-MAST routing-and-aggregation vector
    kernel, excluding common MLLM generation; base MAST precomputes its local
    reconstructions and thus performs no online SVD. MAST-PR shares the same
    offline bank construction and retains additional bases for query-dependent
    reweighting; its vector-kernel latency is not included in the base-MAST
    microbenchmark.}
    \label{tab:efficiency_scaling}
\end{table}

At the default $K{=}5$, task-wise clustering takes $3.780$~s and local SVD
takes $0.081$~s, for $3.930$~s in total. The deployed base-MAST bank stores a
routing context, raw medoid direction, and precomputed local reconstruction
per prototype, requiring $3Cd$ fp32 scalars. MAST-PR additionally retains the
cluster mean and rank-$r$ local basis, requiring approximately $(3+r)Cd$
scalars (with negligible scalar metadata). Thus, bank memory grows linearly
with prototype count, while the local SVD remains an offline operation.

\section{Causal-Alignment Control}
\label{sec:appendix_causal_control}

The main protocol transfers a direction extracted from teacher-forced
completion states to answer-free generation. To test whether this direction can
be replaced by a contrast at a support/query-aligned, answer-free position, we
run a validation-only control on LLaVA-v1.5 at layer~16. For each support
example, neither branch contains an answer. Instead, the positive branch appends
``Carefully ground the answer in the image. Use visual evidence to select the
correct option.'' to the image--question prompt, whereas the negative branch
appends ``Ignore the image and select an option without using visual evidence.''
We extract both states at the final prompt token and construct their difference.
Support routing contexts and evaluation queries use the same positive instruction
and the same answer-free token position. Thus no gold answer, incorrect answer,
or candidate answer enters any feature-extraction or evaluation input.

We retain the frozen MAST configuration ($K{=}5$, $k{=}3$, $r{=}8$,
$T{=}0.5$, and the default shrinkage settings) and run this control only on the
held-out CV-Bench validation split; it is neither used to select hyperparameters
nor substituted for the main test-set protocol. To isolate the effect of the
aligned direction from that of the positive instruction itself, the no-steering
row uses the same positive instruction. Table~\ref{tab:causal_alignment_control}
shows that the answer-free instruction contrast does not retain the benefit of
the completion-conditioned protocol under this fixed configuration. We therefore
retain the explicitly specified teacher-forced completion-to-generation transfer
protocol for all main results. This result is limited to this instruction pair,
model, validation split, and frozen hyperparameter setting; it does not imply
that every possible answer-free contrast is ineffective.

\begin{table}[htbp]
    \centering
    \resizebox{\columnwidth}{!}{%
    \begin{tabular}{lrrrrr}
        \hline
        Method & Relation & Count & Distance & Depth & Average \\
        \hline
        Positive-instruction no-steering & 56.15 & 43.04 & 33.33 & 56.67 & 47.30 \\
        Causal-aligned MAST & 50.77 & 16.46 & 9.17 & 11.67 & 22.01 \\
        \hline
    \end{tabular}
    }
    \caption{Answer-free causal-alignment control on the held-out CV-Bench
    validation split with LLaVA-v1.5 at layer~16. Positive and negative support
    states differ only in their behavior instruction and are both read at the
    final prompt token; routing and query contexts use the positive instruction
    at that same position. Accuracy ($\%$).}
    \label{tab:causal_alignment_control}
\end{table}

\section{Main Results: Additional Analysis}
\label{sec:appendix_main}

\textbf{CV-Bench static-baseline behavior.} On LLaVA-v1.5, MAST's per-task gains
over No-steering are $19.42$, $6.35$, $16.25$, and $26.25$ points on Relation,
Count, Distance, and Depth. Even after each baseline is tuned to its own best
injection strength, no static-direction baseline improves meaningfully over
No-steering: CAA and Act-Add reach only $49.72\%$ and $50.56\%$, both below the
No-steering average of $50.94\%$, and Linear-AcT falls to $44.97\%$. On Qwen2-VL,
raising $\alpha$ quickly degrades and eventually collapses the static baselines,
so per-baseline tuning drives their strength toward near-zero intervention,
leaving every tuned baseline within $0.4$ points of the $77.40\%$ No-steering
average. MAST instead regulates its magnitude through per-cluster shrinkage,
confirming that magnitude adaptivity is necessary to accommodate the differing
activation scales of different models.

\textbf{McNemar significance test.} Table~\ref{tab:mcnemar_significance} reports
the McNemar paired significance test of MAST against the steering baselines on the
CV-Bench test split, summarized in the main-paper Main Results. For each model and
task it lists the least favorable baseline, i.e.\ the one with the largest
$p$-value among the seven steering baselines (CAA, Act-Add, Linear-AcT, ACT, L2S,
Norm-Rnd, RepE), and every other baseline is therefore separated from MAST with a
smaller $p$. Here $b$ counts the discordant samples where MAST is correct and the
baseline is wrong, and $c$ the reverse, so $b>c$ indicates MAST wins the
disagreements. On LLaVA-v1.5 the improvement of MAST over all seven baselines is
significant on all four tasks ($p<0.01$, worst case $p{=}2.0\times10^{-3}$). On
Qwen2-VL it is significant on Relation, Distance, and Depth. The sole exception is
Count, where every method sits within one point of the near-saturated Count
accuracy and MAST, while still ahead of each baseline in the discordant pairs
($b>c$), is not statistically distinguishable ($p{=}0.68$). Correctness follows
the same canonical option-letter matching as the main-paper CV-Bench table, and
paired samples are aligned by index.

\begin{table}[htbp]
    \centering
    {\small
    \setlength{\tabcolsep}{5pt}
    \begin{tabular}{llrrr}
        \hline
        Task & Hardest baseline & $b$ & $c$ & McNemar $p$ \\
        \hline
        \multicolumn{5}{l}{\textit{LLaVA-v1.5}} \\
        Relation & L2S        & 129 & 38 & $<\!10^{-3}$ \\
        Count    & Norm-Rnd   &  49 & 22 & $2.0\times10^{-3}$ \\
        Distance & Linear-AcT & 122 & 72 & $<\!10^{-3}$ \\
        Depth    & RepE       & 119 &  2 & $<\!10^{-3}$ \\
        \hline
        \multicolumn{5}{l}{\textit{Qwen2-VL}} \\
        Relation & Act-Add    &  12 &  0 & $1.5\times10^{-3}$ \\
        Count    & RepE       &  13 & 10 & $0.68$ \\
        Distance & RepE       &  32 & 11 & $2.3\times10^{-3}$ \\
        Depth    & Norm-Rnd   &  41 & 18 & $4.2\times10^{-3}$ \\
        \hline
    \end{tabular}}
    \caption{McNemar's paired significance test of MAST against the steering
    baselines on CV-Bench (test split). For each model and task we report the
    least favorable (largest $p$) of the seven steering baselines, and $b$ and $c$
    are the discordant-pair counts ($b>c$ favors MAST).}
    \label{tab:mcnemar_significance}
\end{table}

\textbf{MAST-PR per-task and monotonicity analysis.} On both models the four-task
average satisfies the monotonic relation MAST $\le$ MAST-PR $\le$ Task-Oracle,
where the Task-Oracle applies the optimal hard direction weights of the known
task. The largest MAST-PR gain on LLaVA-v1.5 lands on Count, the task with the
weakest cross-cluster conflict (Observation~3): Count benefits most when the
fifth local-direction family is emphasized, which the task-conditioned
reweighting of MAST-PR can isolate. Thus, routing supplies locality and the
local basis supplies a compact set of directions for task-specialized weighting.
On Qwen2-VL the soft
routing of MAST-PR slightly exceeds the oracle on Count and Distance, so it is an
upper bound only in the average sense, whereas on LLaVA-v1.5 MAST-PR stays below
the oracle on every task. On LLaVA-v1.5 the oracle exceeds MAST by $16.63$ points,
indicating ample room for direction specialization, whereas on Qwen2-VL it is
only $2.09$ points higher, indicating that at its optimal layer the equal-weight
aggregation of plain MAST is already near optimal.

\textbf{POPE per-split gains.} On LLaVA-v1.5, MAST improves the accuracy of the
Random, Popular, and Adversarial splits by $1.07$, $0.73$, and $0.66$ points
respectively over No-steering.

\section{Qualitative Examples}
\label{sec:appendix_qualitative}

\begin{figure}[htbp]
    \centering
    \includegraphics[width=0.95\linewidth]{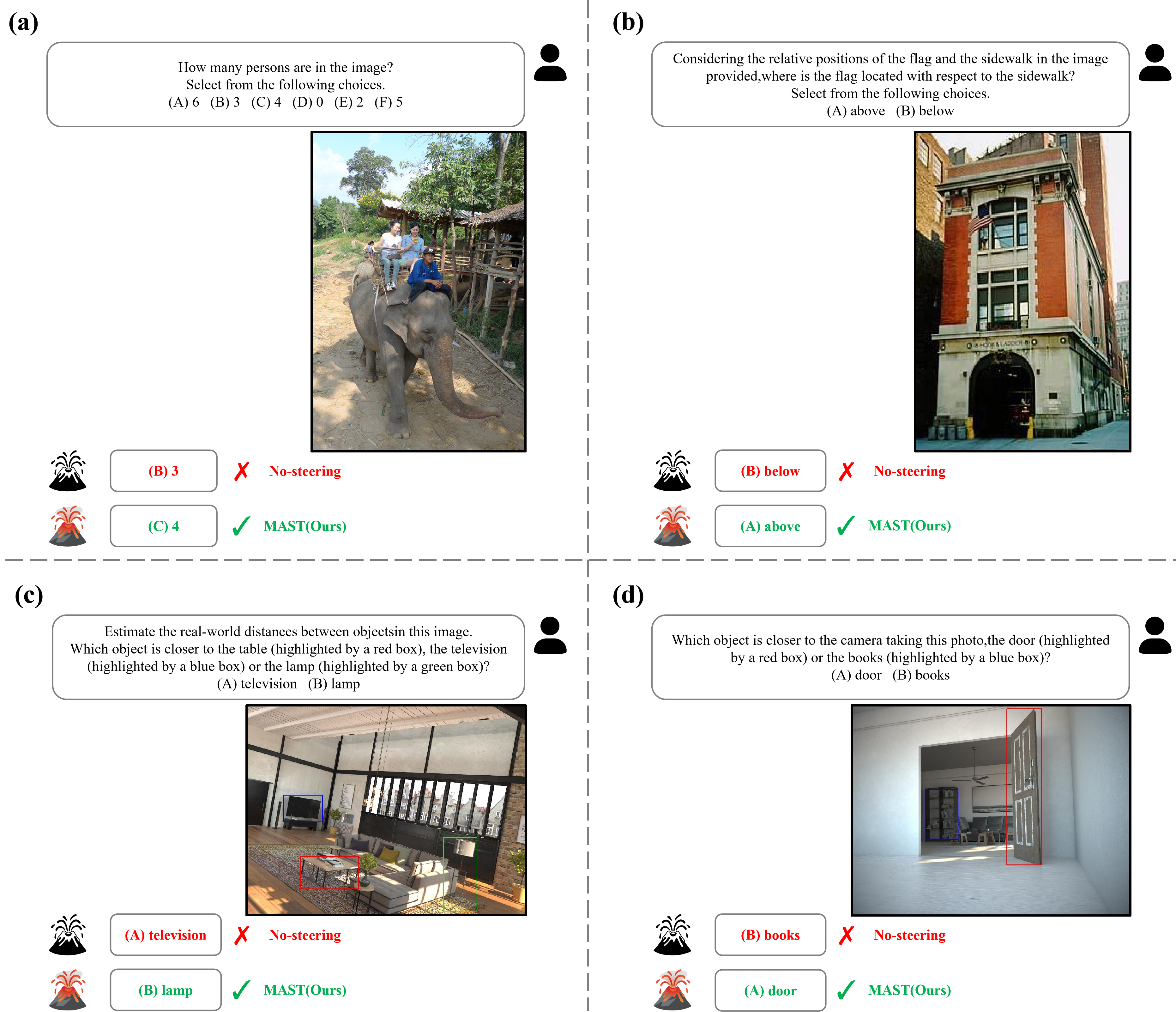}
    \caption{Qualitative examples on the four CV-Bench tasks, one per task.
    They illustrate cases in which MAST uses a routed local steering signal and
    produces the correct answer where the displayed No-steering output is
    incorrect.}
    \label{fig:case_study_2x2}
\end{figure}

Figure~\ref{fig:case_study_2x2} provides one qualitative example per CV-Bench
task, complementing the Count example in the main paper.

\section{Component Ablation: Degeneration Mechanism}
\label{sec:appendix_ablation}

Table~\ref{tab:component_ablation_full} gives the full per-task accuracies of the
leave-one-out component ablation whose four-task average is reported in the main
paper. The per-task columns confirm that the local basis and shrinkage
concentrate almost entirely on the most geometric task, Distance ($1.04$ and
$3.12$ points), while removing multi-center routing collapses every task at once.

\begin{table}[htbp]
    \centering
    \setlength{\tabcolsep}{4pt}
    \begin{tabular}{lcccc rrrrr}
        \hline
        Variant & Md & Mc & LB & Sh & Relation & Count & Distance & Depth & Average \\
        \hline
        \textbf{MAST-PR} & \checkmark & \checkmark & \checkmark & \checkmark & 87.69 & 74.60 & 51.67 & 90.42 & \textbf{76.09} \\
        \hline
        \textbf{MAST} & \checkmark & \checkmark & \checkmark & \checkmark & 79.04 & 53.81 & 39.58 & 99.58 & \textbf{68.00} \\
        \;w/o Shrinkage      & \checkmark & \checkmark & \checkmark & $\times$ & 78.46 & 53.49 & 36.46 & 99.58 & 67.00 \\
        \;w/o Local Basis    & \checkmark & \checkmark & $\times$ & \checkmark & 79.42 & 53.33 & 38.54 & 99.38 & 67.67 \\
        \;w/o Multi-center   & \checkmark & $\times$ & \checkmark & \checkmark & 49.42 & 43.33 & 14.17 & 51.25 & 39.54 \\
        \;w/o Medoid         & $\times$ & $\times$ & $\times$ & $\times$ & 50.19 & 41.90 & 25.42 & 71.88 & 47.35 \\
        \hline
    \end{tabular}
    \caption{Full per-task accuracy ($\%$) of the leave-one-out component
    ablation on CV-Bench (test split). Md is medoid prototype, Mc
    multi-center routing, LB local basis, and Sh adaptive shrinkage. The MAST
    block removes one component at a time under equal weighting.}
    \label{tab:component_ablation_full}
\end{table}

\begin{table}[htbp]
    \centering
    \setlength{\tabcolsep}{5pt}
    \resizebox{\textwidth}{!}{%
    \begin{tabular}{lcccrrrrr}
        \hline
        Subspace representation & Local & Data-driven & Task-cond. & Relation & Count & Distance & Depth & Average \\
        \hline
        Random local basis & \checkmark & $\times$ & \checkmark & $80.05\pm0.44$ & $52.88\pm0.26$ & $40.72\pm0.82$ & $99.17\pm0.00$ & 68.20 \\
        Raw local directions & \checkmark & \checkmark & \checkmark & 90.31 & 55.08 & 55.67 & 98.83 & 74.97 \\
        Global PCA subspace & $\times$ & \checkmark & \checkmark & 90.46 & 54.57 & 66.33 & 99.33 & 77.67 \\
        \textbf{Local PCA/SVD subspace} & \checkmark & \checkmark & \checkmark & \textbf{90.62} & \textbf{72.08} & \textbf{74.17} & \textbf{99.33} & \textbf{84.05} \\
        \hline
    \end{tabular}}
    \caption{Full controlled task-oracle comparison on LLaVA-v1.5 CV-Bench.
    All rows share the MAST backbone, medoid routing, $K{=}5$, $r{=}8$,
    layer, support protocol, and evaluation. For task-conditioned rows,
    ground-truth task labels are used only to select a subspace direction and
    weight on a disjoint validation split, which are frozen before testing.}
    \label{tab:subspace_controls_full}
\end{table}

Table~\ref{tab:subspace_controls_full} complements the equal-weight
leave-one-out ablation by directly testing task-specialized direction families. Under fixed
routing and evaluation, local PCA/SVD subspaces outperform global PCA, raw
local directions, and random local bases, supporting the role of local
subspace directions in task-specialized reweighting.

This section expands on the degenerate variants of the leave-one-out component
ablation in the main paper, explaining why the fully stripped variant behaves as
it does and why the local basis and shrinkage lose their effect once
multi-center routing is removed. The w/o~Medoid variant strips every component
and applies the global mean at the default unit strength. It is therefore not
identical to the separately $\alpha$-tuned CAA baseline of the main-paper
CV-Bench results, which accounts for the small numerical gap between them. When
the number of medoids drops to one, a single real sample tends to lie at the
distribution boundary and is less central than the global mean, and a single
cluster covering all samples makes the intra-cluster residual structure diverge,
so the local basis and the shrinkage lose the local geometry they are meant to
model. This is why the local basis and shrinkage act only as refinement
mechanisms coupled to multi-center routing, effective only when multi-center
routing is already in place, rather than as independently effective components.

\section{Intervention Layer: Per-Task Analysis}
\label{sec:appendix_layer}

This section details the per-task behavior behind the layer scan summarized in
the main paper. Relation together with the two geometric tasks (Distance and
Depth) all attain their strongest gains within layers~11--18, whereas Count
contributes a smaller gain over a broad mid-to-late range. The sole exception is
a sharp Distance-only spike at layer~8, where Distance gains $40.00$ points while
Relation and Count instead degrade. This layer is therefore a single-task
specialization layer for Distance rather than a generally useful one. Under the
unified default layer~16 Distance does not run at its own optimum, so the unified
setting is conservative for individual tasks and leaves room for the
direction-level specialization exploited by MAST-PR.

\section{Hyperparameter Sensitivity: Full Sweep}
\label{sec:appendix_hyperparam}

\begin{figure}[htbp]
    \centering
    \includegraphics[width=0.60\linewidth]{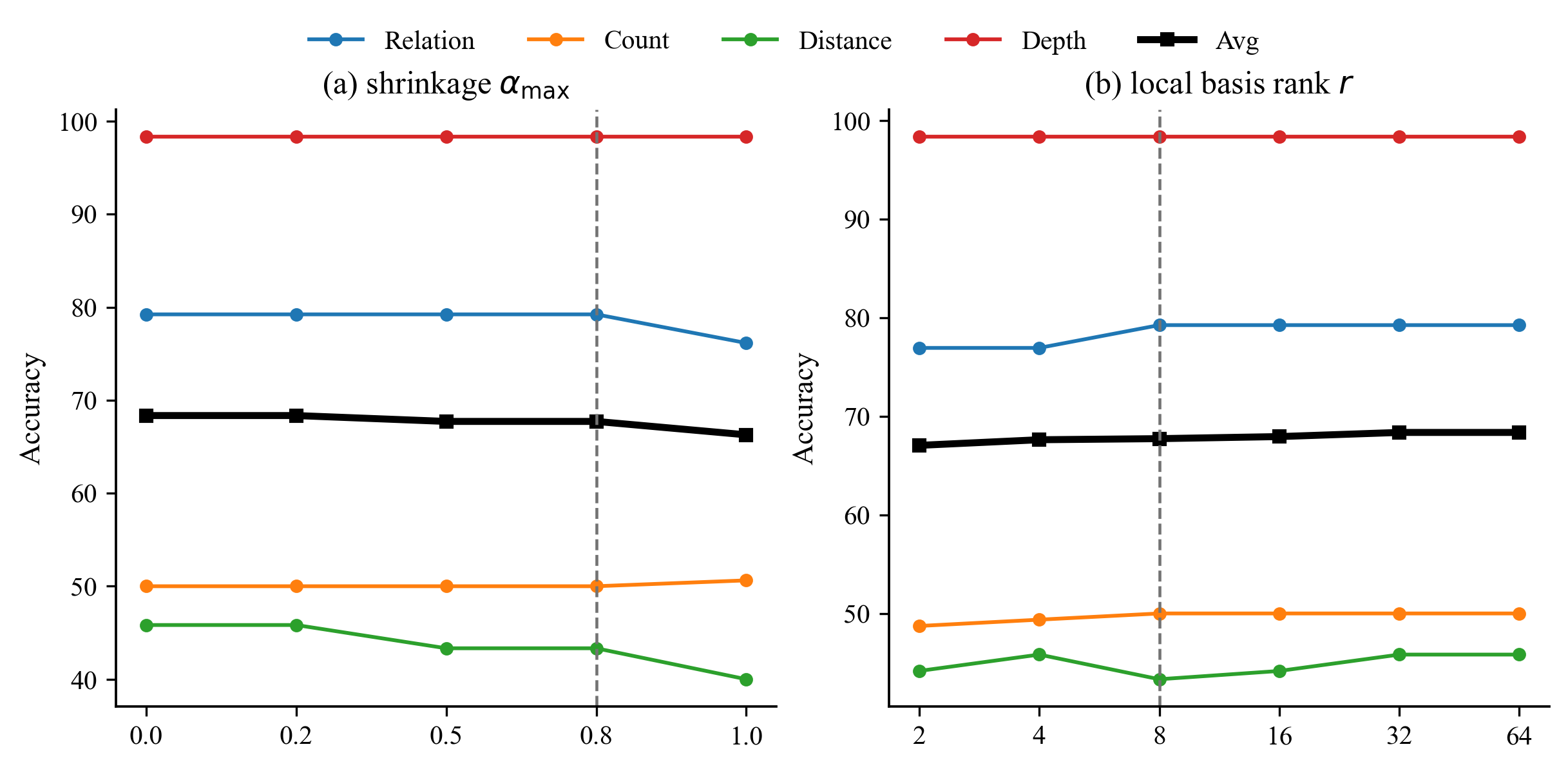}
    \caption{Sensitivity of the shrinkage bound $\alpha_{\max}$ (a) and the local
    basis rank $r$ (b) on CV-Bench (validation split). Thin lines are per-task accuracy and the thick
    line is the four-task average. Sweeping $\alpha_{\max}$ over its full range
    changes the average by only $1.40$ points and sweeping $r$ from~1 to~4096 by
    only $0.76$ points, confirming that the effective steering signal concentrates
    in the leading few singular directions and that MAST is insensitive to both.}
    \label{fig:alpha_r_sensitivity}
\end{figure}

Figure~\ref{fig:alpha_r_sensitivity} reports the sensitivity to the shrinkage
bound $\alpha_{\max}$ and the local basis rank $r$, and
Table~\ref{tab:hyperparam_sensitivity} reports the full one-at-a-time sweep of
the four structural hyperparameters, both on CV-Bench with LLaVA-v1.5 at
layer~16. Sweeping $\alpha_{\max}$ over its full range changes the four-task
average by only $1.40$ points and sweeping $r$ from~1 to~4096 by only $0.76$
points, so the effective steering signal concentrates in the leading few singular
directions and MAST is insensitive to both. For $K$ and $k$ the average is
maximized at the extremes ($K{=}64$, $k{=}1$) only because Relation and Count
saturate there, whereas the geometric task Distance collapses ($20.83\%$ at
$K{=}64$, $26.67\%$ at $k{=}1$), so the moderate defaults $K{=}5$ and $k{=}3$ are
chosen to preserve Distance.

\begin{table}[htbp]
    \centering
    \setlength{\tabcolsep}{2.5pt}
    \begin{tabular}{lrrrrrr}
        \hline
        Param & Value & Relation & Count & Distance & Depth & Average \\
        \hline
        \multirow{8}{*}{$K$}
          & 1            & 50.77 & 38.61 & 10.00 & 44.17 & 35.89 \\
          & 2            & 50.77 & 44.30 & 18.33 & 90.83 & 51.06 \\
          & 4            & 74.62 & 49.37 & 27.50 & 99.17 & 62.66 \\
          & \textbf{5}   & \textbf{79.23} & \textbf{50.00} & \textbf{43.33} & \textbf{98.33} & \textbf{67.72} \\
          & 8            & 74.62 & 54.43 & 35.83 & 100.00 & 66.22 \\
          & 16           & 95.38 & 61.39 & 33.33 & 100.00 & 72.53 \\
          & 32           & 100.00 & 68.99 & 22.50 & 99.17 & 72.66 \\
          & 64           & 100.00 & 71.52 & 20.83 & 99.17 & 72.88 \\
        \hline
        \multirow{8}{*}{$r$}
          & 1            & 82.31 & 49.37 & 40.00 & 98.33 & 67.50 \\
          & 2            & 76.92 & 48.73 & 44.17 & 98.33 & 67.04 \\
          & 4            & 76.92 & 49.37 & 45.83 & 98.33 & 67.61 \\
          & \textbf{8}   & \textbf{79.23} & \textbf{50.00} & \textbf{43.33} & \textbf{98.33} & \textbf{67.72} \\
          & 16           & 79.23 & 50.00 & 44.17 & 98.33 & 67.93 \\
          & 32           & 79.23 & 50.00 & 45.83 & 98.33 & 68.35 \\
          & 64           & 79.23 & 50.00 & 45.83 & 98.33 & 68.35 \\
          & 4096         & 79.23 & 50.00 & 45.83 & 98.33 & 68.35 \\
        \hline
        \multirow{5}{*}{$k$}
          & 1            & 98.46 & 63.92 & 26.67 & 100.00 & 72.26 \\
          & 2            & 93.08 & 53.80 & 34.17 & 99.17 & 70.05 \\
          & \textbf{3}   & \textbf{79.23} & \textbf{50.00} & \textbf{43.33} & \textbf{98.33} & \textbf{67.72} \\
          & 4            & 68.46 & 47.47 & 37.50 & 97.50 & 62.73 \\
          & 8            & 60.00 & 47.47 & 32.50 & 95.00 & 58.74 \\
        \hline
        \multirow{5}{*}{$\alpha_{\max}$}
          & 0.0          & 79.23 & 50.00 & 45.83 & 98.33 & 68.35 \\
          & 0.2          & 79.23 & 50.00 & 45.83 & 98.33 & 68.35 \\
          & 0.5          & 79.23 & 50.00 & 43.33 & 98.33 & 67.72 \\
          & \textbf{0.8} & \textbf{79.23} & \textbf{50.00} & \textbf{43.33} & \textbf{98.33} & \textbf{67.72} \\
          & 1.0          & 76.15 & 50.63 & 40.00 & 98.33 & 66.28 \\
        \hline
    \end{tabular}
    \caption{Full one-at-a-time hyperparameter sweep on CV-Bench (validation
    split). Each block varies one parameter with the others at their defaults
    (bold). $K$ is the number of medoids, $r$ the local basis rank, $k$ the routing
    size, and $\alpha_{\max}$ the upper shrinkage bound. The average is nearly flat in
    $r$ and $\alpha_{\max}$, confirming that MAST is insensitive to both. For $K$
    and $k$ the average is maximized at the extremes ($K{=}64$, $k{=}1$), but only
    because Relation and Count saturate: the geometric task Distance instead
    collapses there ($20.83$ at $K{=}64$, $26.67$ at $k{=}1$), so the moderate
    defaults $K{=}5$ and $k{=}3$ are chosen to preserve Distance rather than the
    average-maximizing corner.}
    \label{tab:hyperparam_sensitivity}
\end{table}

\section{Direction Specialization: Full Weight \texorpdfstring{$\times$}{x} Direction Sweep}
\label{sec:appendix_direction}

This section reports the complete weight $\times$ direction-family-rank sweep behind the
condensed direction-specialization table of the main paper
(Tables~\ref{tab:dirscan_relation}--\ref{tab:dirscan_depth_qwen}), for both models
(LLaVA-v1.5 at layer~16 and Qwen2-VL at layer~4), on CV-Bench (validation
split). For each CV-Bench task we emphasize the $d$-th local-basis direction in
every cluster, i.e., the direction family $\{b_{c,d}\}_{c\in\mathcal{C}}$ for
$d\in\{1,\dots,8\}$, with weight $w$ under sv-norm scaling and zero the remaining
directions, and report the resulting validation accuracy. The equal-weight
aggregation used by MAST (\emph{Equal}) is listed per task as a reference. For each
task the single (direction-family rank, weight) cell that maximizes accuracy (\textbf{bold})
is the one frozen as that task's MAST-PR / Task-Oracle weight in the main paper.
The preferred direction families differ across the two models: on LLaVA-v1.5,
Relation, Count, and Distance favor Dir-6, Dir-5, and Dir-2 while Depth keeps equal
weighting. On Qwen2-VL they favor Dir-6, Dir-8, Dir-8, and Dir-6 respectively,
which is exactly why MAST-PR uses per-model direction-family weights. These are
operational performance preferences over cluster-local coordinates, not a claim
that a task-pure cluster contains directions of multiple task identities. The
tables also make the geometry concrete: accuracy collapses as $w$ grows because an
over-amplified direction family dominates the steering vector, and each
task-specialized family attains its peak at a moderate weight before this
collapse.

\begin{table}[htbp]
    \centering
    \setlength{\tabcolsep}{3.2pt}
    \begin{tabular}{lrrrrrrrr}
        \hline
        $w$ & Dir-1 & Dir-2 & Dir-3 & Dir-4 & Dir-5 & Dir-6 & Dir-7 & Dir-8 \\
        \hline
        1   & 82.31 & 75.38 & 76.15 & 76.92 & 81.54 & 78.46 & 79.23 & 76.92 \\
        5   & 81.54 & 66.92 & 67.69 & 80.77 & 85.38 & 83.85 & 82.31 & 80.77 \\
        10  & 83.08 & 62.31 & 56.15 & 81.54 & 84.62 & 84.62 & 81.54 & 83.85 \\
        20  & 69.23 & 76.15 & 53.08 & 82.31 & 61.54 & 88.46 & 72.31 & 84.62 \\
        50  & 50.00 & 40.00 & 36.92 & 43.85 &  1.54 & \textbf{93.08} & 32.31 & 68.46 \\
        75  & 42.31 & 26.15 &  6.15 & 26.92 &  0.00 & 49.23 & 21.54 & 20.00 \\
        100 & 20.77 & 25.38 &  6.15 & 24.62 &  0.00 & 39.23 & 20.00 &  9.23 \\
        125 &  2.31 & 16.92 &  0.77 & 15.38 &  0.00 & 36.92 &  0.00 &  1.54 \\
        150 &  0.77 &  3.85 &  0.00 &  0.77 &  0.00 &  2.31 &  0.00 &  2.31 \\
        175 &  0.00 &  0.00 &  0.77 &  0.00 &  0.00 &  2.31 &  0.00 &  0.77 \\
        200 &  0.00 &  0.00 &  0.00 &  0.00 &  0.00 &  0.77 &  0.00 &  0.77 \\
        \hline
    \end{tabular}
    \caption{LLaVA-v1.5, Relation direction-specialization sweep (Equal $=79.23$,
    No-steering $=54.62$). Selected weight: Dir-6 at $w{=}50$ (bold).}
    \label{tab:dirscan_relation}
\end{table}

\begin{table}[htbp]
    \centering
    \setlength{\tabcolsep}{3.2pt}
    \begin{tabular}{lrrrrrrrr}
        \hline
        $w$ & Dir-1 & Dir-2 & Dir-3 & Dir-4 & Dir-5 & Dir-6 & Dir-7 & Dir-8 \\
        \hline
        1   & 49.37 & 49.37 & 50.63 & 50.63 & 51.90 & 50.00 & 51.27 & 50.63 \\
        5   & 48.10 & 50.00 & 53.80 & 50.63 & 55.70 & 50.00 & 53.80 & 51.27 \\
        10  & 44.30 & 51.27 & 54.43 & 51.27 & 64.56 & 49.37 & 58.23 & 52.53 \\
        20  & 43.67 & 46.84 & 60.76 & 49.37 & \textbf{72.15} & 50.00 & 63.29 & 51.90 \\
        50  & 34.18 & 43.04 & 31.01 & 37.97 & 44.94 & 34.81 & 68.99 & 22.78 \\
        75  & 24.05 & 39.24 & 22.78 & 23.42 &  9.49 & 20.89 & 53.80 & 20.25 \\
        100 & 10.76 & 33.54 & 24.05 & 17.72 &  1.90 & 16.46 & 31.65 & 15.19 \\
        125 &  3.80 & 25.32 & 21.52 &  7.59 &  0.00 & 11.39 & 24.68 & 17.09 \\
        150 &  2.53 & 23.42 & 26.58 &  4.43 &  0.00 &  5.06 & 13.29 & 14.56 \\
        175 &  5.06 & 12.66 & 36.08 &  4.43 &  0.00 &  5.70 & 18.35 &  9.49 \\
        200 &  0.00 &  4.43 & 37.97 &  3.80 &  0.00 &  4.43 & 20.25 &  9.49 \\
        \hline
    \end{tabular}
    \caption{LLaVA-v1.5, Count direction-specialization sweep (Equal $=50.00$,
    No-steering $=46.84$). Selected weight: Dir-5 at $w{=}20$ (bold).}
    \label{tab:dirscan_count}
\end{table}

\begin{table}[htbp]
    \centering
    \setlength{\tabcolsep}{3.2pt}
    \begin{tabular}{lrrrrrrrr}
        \hline
        $w$ & Dir-1 & Dir-2 & Dir-3 & Dir-4 & Dir-5 & Dir-6 & Dir-7 & Dir-8 \\
        \hline
        1   & 40.00 & 43.33 & 40.83 & 42.50 & 44.17 & 41.67 & 40.00 & 38.33 \\
        5   & 43.33 & 48.33 & 39.17 & 47.50 & 40.00 & 32.50 & 36.67 & 30.00 \\
        10  & 35.00 & 46.67 & 35.00 & 45.00 & 25.00 & 17.50 & 27.50 & 19.17 \\
        20  & 34.17 & 40.83 & 37.50 & 41.67 & 11.67 & 19.17 & 20.00 & 13.33 \\
        50  & 26.67 & 45.00 & 22.50 & 44.17 & 25.83 & 35.83 & 10.83 & 11.67 \\
        75  & 40.00 & 69.17 & 13.33 & 19.17 &  0.00 & 30.83 &  1.67 & 27.50 \\
        100 & 39.17 & \textbf{74.17} & 18.33 & 15.83 &  0.00 & 33.33 &  5.00 & 25.83 \\
        125 & 51.67 & 67.50 & 10.00 & 25.83 &  0.00 & 32.50 &  8.33 & 28.33 \\
        150 & 55.00 & 56.67 & 10.00 & 31.67 &  0.00 & 30.00 &  2.50 & 11.67 \\
        175 & 38.33 & 55.00 & 10.83 & 10.83 &  0.00 &  1.67 &  0.00 &  0.00 \\
        200 &  0.00 & 55.00 &  5.83 &  0.00 &  0.00 &  5.00 &  0.00 &  0.83 \\
        \hline
    \end{tabular}
    \caption{LLaVA-v1.5, Distance direction-specialization sweep (Equal $=43.33$,
    No-steering $=26.67$). Selected weight: Dir-2 at $w{=}100$ (bold).}
    \label{tab:dirscan_distance}
\end{table}

\begin{table}[htbp]
    \centering
    \setlength{\tabcolsep}{3.2pt}
    \begin{tabular}{lrrrrrrrr}
        \hline
        $w$ & Dir-1 & Dir-2 & Dir-3 & Dir-4 & Dir-5 & Dir-6 & Dir-7 & Dir-8 \\
        \hline
        1   & 98.33 & 98.33 & 98.33 & 98.33 & 98.33 & 98.33 & 98.33 & 98.33 \\
        5   & 98.33 & 96.67 & 98.33 & 99.17 & 98.33 & 99.17 & 98.33 & 97.50 \\
        10  & 99.17 & 95.83 & 98.33 & 100.00 & 99.17 & 100.00 & 95.00 & 95.83 \\
        20  & 99.17 & 99.17 & 96.67 & 62.50 & 99.17 & 100.00 & 76.67 & 93.33 \\
        50  & 50.00 & 89.17 & 89.17 &  1.67 & 43.33 & 90.00 & 14.17 & 63.33 \\
        75  &  5.00 & 39.17 & 83.33 &  3.33 &  9.17 & 81.67 & 12.50 & 77.50 \\
        100 & 13.33 & 15.00 & 53.33 &  0.00 &  0.83 & 44.17 &  6.67 & 66.67 \\
        125 & 17.50 &  5.00 & 19.17 &  0.00 &  0.00 & 11.67 &  3.33 & 40.83 \\
        150 &  4.17 &  0.83 &  0.00 &  0.00 &  0.00 &  2.50 &  0.00 & 23.33 \\
        175 &  0.00 &  0.00 &  0.00 &  0.00 &  0.00 &  5.00 &  0.00 & 19.17 \\
        200 &  0.00 &  0.00 &  0.00 &  0.00 &  0.00 &  1.67 &  0.00 &  9.17 \\
        \hline
    \end{tabular}
    \caption{LLaVA-v1.5, Depth direction-specialization sweep (Equal $=98.33$,
    No-steering $=75.83$). No single direction meaningfully beats Equal, so Depth
    retains equal weighting ($\beta^{\text{Depth}}$ all-ones).}
    \label{tab:dirscan_depth}
\end{table}

\begin{table}[htbp]
    \centering
    \setlength{\tabcolsep}{3.2pt}
    \begin{tabular}{lrrrrrrrr}
        \hline
        $w$ & Dir-1 & Dir-2 & Dir-3 & Dir-4 & Dir-5 & Dir-6 & Dir-7 & Dir-8 \\
        \hline
        1   & 80.00 & 80.77 & 81.54 & 81.54 & 80.77 & 81.54 & 81.54 & 81.54 \\
        5   & 79.23 & 80.77 & 80.00 & 81.54 & 80.77 & 81.54 & 80.77 & 81.54 \\
        10  & 76.92 & 80.00 & 79.23 & 81.54 & 80.77 & 82.31 & 80.00 & 81.54 \\
        20  & 26.15 & 80.77 & 78.46 & 83.08 & 80.77 & \textbf{83.85} & 80.00 & 80.77 \\
        50  &  4.62 & 33.85 & 43.08 & 79.23 & 79.23 & 50.77 & 59.23 & 81.54 \\
        75  &  3.08 & 23.85 & 45.38 & 27.69 & 80.00 & 33.08 & 13.85 & 80.77 \\
        100 &  2.31 & 19.23 & 23.08 & 24.62 & 81.54 & 38.46 &  7.69 & 80.77 \\
        125 &  2.31 & 16.92 & 10.77 & 14.62 & 81.54 & 13.85 &  6.92 & 80.77 \\
        150 &  2.31 & 12.31 &  8.46 &  5.38 & 81.54 & 33.08 &  6.92 & 80.77 \\
        175 &  2.31 & 13.85 &  3.85 &  6.92 & 83.08 & 13.08 &  6.92 & 80.77 \\
        200 &  2.31 & 10.77 &  3.08 & 14.62 & 83.08 &  6.15 &  4.62 & 80.77 \\
        \hline
    \end{tabular}
    \caption{Qwen2-VL, Relation direction-specialization sweep (Equal $=80.00$).
    Selected weight: Dir-6 at $w{=}20$ (bold).}
    \label{tab:dirscan_relation_qwen}
\end{table}

\begin{table}[htbp]
    \centering
    \setlength{\tabcolsep}{3.2pt}
    \begin{tabular}{lrrrrrrrr}
        \hline
        $w$ & Dir-1 & Dir-2 & Dir-3 & Dir-4 & Dir-5 & Dir-6 & Dir-7 & Dir-8 \\
        \hline
        1   & 68.35 & 69.62 & 68.35 & 68.99 & 68.99 & 68.35 & 68.99 & \textbf{70.25} \\
        5   & 63.92 & 62.66 & 68.99 & 68.99 & 69.62 & 67.72 & 65.82 & 65.19 \\
        10  & 50.63 & 51.90 & 69.62 & 68.99 & 69.62 & 54.43 & 42.41 & 49.37 \\
        20  & 32.91 & 30.38 & 65.82 & 68.35 & 66.46 &  1.27 & 12.66 & 16.46 \\
        50  & 15.82 & 15.19 & 39.87 & 68.99 & 43.04 &  0.00 &  5.06 & 21.52 \\
        75  & 18.99 &  6.96 & 22.15 & 65.19 & 11.39 &  0.00 &  0.00 & 24.05 \\
        100 & 15.19 &  5.06 &  8.23 & 41.77 &  6.33 &  0.00 &  0.00 & 18.35 \\
        125 & 11.39 &  2.53 & 13.92 & 31.65 &  3.80 &  0.00 &  0.00 & 12.03 \\
        150 & 10.13 &  4.43 &  6.96 & 26.58 &  1.27 &  0.00 &  0.00 &  6.96 \\
        175 & 12.66 &  3.80 &  7.59 & 26.58 &  0.00 &  0.00 &  0.00 &  0.63 \\
        200 &  5.06 &  0.00 &  6.33 & 24.68 &  0.00 &  0.00 &  0.00 &  0.00 \\
        \hline
    \end{tabular}
    \caption{Qwen2-VL, Count direction-specialization sweep (Equal $=68.99$).
    Selected weight: Dir-8 at $w{=}1$ (bold).}
    \label{tab:dirscan_count_qwen}
\end{table}

\begin{table}[htbp]
    \centering
    \setlength{\tabcolsep}{3.2pt}
    \begin{tabular}{lrrrrrrrr}
        \hline
        $w$ & Dir-1 & Dir-2 & Dir-3 & Dir-4 & Dir-5 & Dir-6 & Dir-7 & Dir-8 \\
        \hline
        1   & 84.17 & 84.17 & 83.33 & 83.33 & 84.17 & 84.17 & 82.50 & 85.83 \\
        5   & 55.00 & 83.33 & 82.50 & 78.33 & 84.17 & 79.17 & 75.00 & \textbf{87.50} \\
        10  & 55.00 & 46.67 & 81.67 & 72.50 & 84.17 & 73.33 & 76.67 & 73.33 \\
        20  & 10.00 &  0.00 & 81.67 & 62.50 & 82.50 & 55.00 & 55.00 & 15.00 \\
        50  &  0.83 &  0.00 & 40.83 & 57.50 &  0.00 &  0.00 & 55.00 & 40.83 \\
        75  &  0.00 &  0.00 &  3.33 & 40.00 &  0.00 &  0.00 &  0.83 &  0.83 \\
        100 &  0.00 &  0.00 &  0.00 &  0.00 &  0.00 &  0.00 &  0.00 &  0.00 \\
        125 &  0.00 &  0.00 &  5.83 &  0.00 &  0.00 &  0.00 &  0.00 &  0.00 \\
        150 &  0.00 &  0.00 &  0.00 &  3.33 &  0.00 &  0.00 &  0.00 &  0.00 \\
        175 &  0.00 &  0.00 &  0.00 &  2.50 &  0.00 &  0.00 &  0.00 &  0.00 \\
        200 &  0.00 &  0.00 &  0.00 & 30.83 &  0.00 &  0.00 &  0.00 &  0.00 \\
        \hline
    \end{tabular}
    \caption{Qwen2-VL, Distance direction-specialization sweep (Equal $=84.17$).
    Selected weight: Dir-8 at $w{=}5$ (bold).}
    \label{tab:dirscan_distance_qwen}
\end{table}

\begin{table}[htbp]
    \centering
    \setlength{\tabcolsep}{3.2pt}
    \begin{tabular}{lrrrrrrrr}
        \hline
        $w$ & Dir-1 & Dir-2 & Dir-3 & Dir-4 & Dir-5 & Dir-6 & Dir-7 & Dir-8 \\
        \hline
        1   & 91.67 & 91.67 & 91.67 & 91.67 & 91.67 & 92.50 & 91.67 & 91.67 \\
        5   & 83.33 & 92.50 & 91.67 & 92.50 & 92.50 & 92.50 & 91.67 & 91.67 \\
        10  & 66.67 & 87.50 & 91.67 & 90.00 & 89.17 & \textbf{94.17} & 91.67 & 93.33 \\
        20  & 66.67 & 59.17 & 85.00 & 52.50 & 47.50 & 90.00 & 91.67 & 75.00 \\
        50  & 62.50 &  4.17 & 25.00 & 15.83 & 51.67 & 34.17 & 84.17 &  1.67 \\
        75  & 55.00 &  5.00 & 21.67 &  0.00 & 41.67 & 15.00 & 79.17 &  0.83 \\
        100 & 10.83 & 54.17 & 33.33 &  0.00 & 18.33 &  0.00 & 85.00 &  0.00 \\
        125 &  6.67 & 55.00 & 34.17 &  0.00 &  1.67 &  0.00 & 80.00 &  0.00 \\
        150 &  7.50 & 53.33 & 32.50 &  0.00 &  1.67 &  0.00 & 77.50 &  0.00 \\
        175 &  8.33 & 20.00 & 18.33 &  0.00 &  1.67 &  0.00 & 80.00 &  0.00 \\
        200 &  9.17 &  0.83 &  0.83 &  0.00 &  1.67 &  0.00 & 78.33 &  0.00 \\
        \hline
    \end{tabular}
    \caption{Qwen2-VL, Depth direction-specialization sweep (Equal $=91.67$).
    Selected weight: Dir-6 at $w{=}10$ (bold).}
    \label{tab:dirscan_depth_qwen}
\end{table}

%% file: MAST_ICLR2027.bbl
\begin{thebibliography}{29}
\providecommand{\natexlab}[1]{#1}
\providecommand{\url}[1]{\texttt{#1}}
\expandafter\ifx\csname urlstyle\endcsname\relax
  \providecommand{\doi}[1]{doi: #1}\else
  \providecommand{\doi}{doi: \begingroup \urlstyle{rm}\Url}\fi

\bibitem[Chalnev et~al.(2024)Chalnev, Siu, and
  Conmy]{chalnev2024improvingsteeringvectorstargeting}
Sviatoslav Chalnev, Matthew Siu, and Arthur Conmy.
\newblock Improving steering vectors by targeting sparse autoencoder features,
  2024.
\newblock URL \url{https://arxiv.org/abs/2411.02193}.

\bibitem[Feng et~al.(2026)Feng, Li, Zhu, Zhou, Qian, Zhang, Deryl, Onn, Ng, and
  Mao]{feng2026finegrained}
Zijian Feng, Tianjiao Li, Zixiao Zhu, Hanzhang Zhou, Junlang Qian, Li~Zhang,
  Chua Jia~Jim Deryl, Mak~Lee Onn, Gee~Wah Ng, and Kezhi Mao.
\newblock Fine-grained activation steering: Steering less, achieving more.
\newblock In \emph{The Fourteenth International Conference on Learning
  Representations}, 2026.
\newblock URL \url{https://openreview.net/forum?id=guSVafqhrB}.

\bibitem[Gan et~al.(2025)Gan, Fu, Asilis, Liu, Yogatama, Sharan, Jia, and
  Neiswanger]{gan2025textualsteeringvectorsimprove}
Woody~Haosheng Gan, Deqing Fu, Julian Asilis, Ollie Liu, Dani Yogatama, Vatsal
  Sharan, Robin Jia, and Willie Neiswanger.
\newblock Textual steering vectors can improve visual understanding in
  multimodal large language models, 2025.
\newblock URL \url{https://arxiv.org/abs/2505.14071}.

\bibitem[Huang et~al.(2026)Huang, Li, Surana, Yu, Wang, McAuley, Shang, and
  Wu]{huang2026ampsadaptivemodalitypreference}
Zihan Huang, Xintong Li, Rohan Surana, Tong Yu, Rui Wang, Julian McAuley,
  Jingbo Shang, and Junda Wu.
\newblock Amps: Adaptive modality preference steering via functional entropy,
  2026.
\newblock URL \url{https://arxiv.org/abs/2602.12533}.

\bibitem[Jang et~al.(2026)Jang, Seo, Cho, Ko, and
  Kim]{jang2026tvsstesttimevisualsubspace}
Jaehyuk Jang, Minseok Seo, Seungju Cho, Kangwook Ko, and Changick Kim.
\newblock T-vss: Test-time visual subspace steering for adversarial robustness
  of vision-language models, 2026.
\newblock URL \url{https://arxiv.org/abs/2606.23132}.

\bibitem[Kang \& Kim(2026)Kang and
  Kim]{kang2026enhancinginstructionfollowingllms}
Minjae Kang and Jaehyung Kim.
\newblock Enhancing instruction following of llms via activation steering with
  dynamic rejection, 2026.
\newblock URL \url{https://arxiv.org/abs/2603.06745}.

\bibitem[Kayan \& Zhang(2025)Kayan and
  Zhang]{kayan2025prototypebaseddynamicsteeringlarge}
Ceyhun~Efe Kayan and Li~Zhang.
\newblock Prototype-based dynamic steering for large language models, 2025.
\newblock URL \url{https://arxiv.org/abs/2510.05498}.

\bibitem[Khayatan et~al.(2025)Khayatan, Shukor, Parekh, Dapogny, and
  Cord]{11445319}
Pegah Khayatan, Mustafa Shukor, Jayneel Parekh, Arnaud Dapogny, and Matthieu
  Cord.
\newblock Analyzing fine-tuning representation shift for multimodal llms
  steering.
\newblock In \emph{2025 IEEE/CVF International Conference on Computer Vision
  (ICCV)}, pp.\  2206--2216, 2025.
\newblock \doi{10.1109/ICCV51701.2025.00213}.

\bibitem[Li et~al.(2026)Li, Liu, Li, Liu, Wang, Liang, Yao, Chang, and
  Xing]{li2026suppressingpriorcomparisonhallucinationsradiology}
Ao~Li, Rui Liu, Mingjie Li, Sheng Liu, Lei Wang, Xiaodan Liang, Lina Yao,
  Xiaojun Chang, and Lei Xing.
\newblock Suppressing prior-comparison hallucinations in radiology report
  generation via semantically decoupled latent steering, 2026.
\newblock URL \url{https://arxiv.org/abs/2602.23676}.

\bibitem[Li et~al.(2023)Li, Du, Zhou, Wang, Zhao, and Wen]{li2023pope}
Yifan Li, Yifan Du, Kun Zhou, Jinpeng Wang, Wayne~Xin Zhao, and Ji-Rong Wen.
\newblock Evaluating object hallucination in large vision-language models.
\newblock In \emph{Proceedings of the 2023 Conference on Empirical Methods in
  Natural Language Processing (EMNLP)}, 2023.

\bibitem[Lin et~al.(2014)Lin, Maire, Belongie, Hays, Perona, Ramanan,
  Doll{\'a}r, and Zitnick]{10.1007/978-3-319-10602-1_48}
Tsung-Yi Lin, Michael Maire, Serge Belongie, James Hays, Pietro Perona, Deva
  Ramanan, Piotr Doll{\'a}r, and C.~Lawrence Zitnick.
\newblock Microsoft coco: Common objects in context.
\newblock In David Fleet, Tomas Pajdla, Bernt Schiele, and Tinne Tuytelaars
  (eds.), \emph{Computer Vision -- ECCV 2014}, pp.\  740--755, Cham, 2014.
  Springer International Publishing.
\newblock ISBN 978-3-319-10602-1.

\bibitem[Liu et~al.(2024)Liu, Li, Li, and Lee]{10655294}
Haotian Liu, Chunyuan Li, Yuheng Li, and Yong~Jae Lee.
\newblock Improved baselines with visual instruction tuning.
\newblock In \emph{2024 IEEE/CVF Conference on Computer Vision and Pattern
  Recognition (CVPR)}, pp.\  26286--26296, 2024.
\newblock \doi{10.1109/CVPR52733.2024.02484}.

\bibitem[Liu et~al.(2025)Liu, Ye, and Zou]{ICLR2025_b4008025}
Sheng Liu, Haotian Ye, and James~Y Zou.
\newblock Reducing hallucinations in large vision-language models via latent
  space steering.
\newblock In Y.~Yue, A.~Garg, N.~Peng, F.~Sha, and R.~Yu (eds.),
  \emph{International Conference on Learning Representations}, volume 2025,
  pp.\  72402--72419, 2025.
\newblock URL
  \url{https://proceedings.iclr.cc/paper_files/paper/2025/file/b4008025c2182bfe16fcc8566ee14d64-Paper-Conference.pdf}.

\bibitem[McInnes et~al.(2020)McInnes, Healy, and
  Melville]{mcinnes2020umapuniformmanifoldapproximation}
Leland McInnes, John Healy, and James Melville.
\newblock Umap: Uniform manifold approximation and projection for dimension
  reduction, 2020.
\newblock URL \url{https://arxiv.org/abs/1802.03426}.

\bibitem[Parekh et~al.(2025)Parekh, Khayatan, Shukor, Dapogny, Newson, and
  Cord]{parekh2025neurips-learning}
Jayneel Parekh, Pegah Khayatan, Mustafa Shukor, Arnaud Dapogny, Alasdair
  Newson, and Matthieu Cord.
\newblock {Learning to Steer: Input-Dependent Steering for Multimodal LLMs}.
\newblock In \emph{Advances in Neural Information Processing Systems}, 2025.
\newblock URL
  \url{https://mlanthology.org/neurips/2025/parekh2025neurips-learning/}.

\bibitem[Park \& Jun(2009)Park and Jun]{PARK20093336}
Hae-Sang Park and Chi-Hyuck Jun.
\newblock A simple and fast algorithm for k-medoids clustering.
\newblock \emph{Expert Systems with Applications}, 36\penalty0 (2, Part
  2):\penalty0 3336--3341, 2009.
\newblock ISSN 0957-4174.
\newblock \doi{https://doi.org/10.1016/j.eswa.2008.01.039}.
\newblock URL
  \url{https://www.sciencedirect.com/science/article/pii/S095741740800081X}.

\bibitem[Phute \& Balakrishnan(2025)Phute and
  Balakrishnan]{phute2025visorvisualinputbasedsteering}
Mansi Phute and Ravikumar Balakrishnan.
\newblock Visor: Visual input-based steering for output redirection in
  vision-language models, 2025.
\newblock URL \url{https://arxiv.org/abs/2508.08521}.

\bibitem[Rimsky et~al.(2024)Rimsky, Gabrieli, Schulz, Tong, Hubinger, and
  Turner]{rimsky-etal-2024-steering}
Nina Rimsky, Nick Gabrieli, Julian Schulz, Meg Tong, Evan Hubinger, and
  Alexander Turner.
\newblock Steering llama 2 via contrastive activation addition.
\newblock In Lun-Wei Ku, Andre Martins, and Vivek Srikumar (eds.),
  \emph{Proceedings of the 62nd Annual Meeting of the Association for
  Computational Linguistics (Volume 1: Long Papers)}, pp.\  15504--15522,
  Bangkok, Thailand, August 2024. Association for Computational Linguistics.
\newblock \doi{10.18653/v1/2024.acl-long.828}.
\newblock URL \url{https://aclanthology.org/2024.acl-long.828/}.

\bibitem[Rodriguez et~al.(2025)Rodriguez, Blaas, Klein, Zappella, Apostoloff,
  marco cuturi, and Suau]{rodriguez2025controlling}
Pau Rodriguez, Arno Blaas, Michal Klein, Luca Zappella, Nicholas Apostoloff,
  marco cuturi, and Xavier Suau.
\newblock Controlling language and diffusion models by transporting
  activations.
\newblock In \emph{The Thirteenth International Conference on Learning
  Representations}, 2025.
\newblock URL \url{https://openreview.net/forum?id=l2zFn6TIQi}.

\bibitem[Stolfo et~al.(2025)Stolfo, Balachandran, Yousefi, Horvitz, and
  Nushi]{stolfo2025improving}
Alessandro Stolfo, Vidhisha Balachandran, Safoora Yousefi, Eric Horvitz, and
  Besmira Nushi.
\newblock Improving instruction-following in language models through activation
  steering.
\newblock In \emph{The Thirteenth International Conference on Learning
  Representations}, 2025.
\newblock URL \url{https://openreview.net/forum?id=wozhdnRCtw}.

\bibitem[Tong et~al.(2024)Tong, Brown, Wu, Woo, Middepogu, Akula, Yang, Yang,
  Iyer, Pan, Wang, Fergus, LeCun, and Xie]{tong2024cambrian}
Shengbang Tong, Ellis Brown, Penghao Wu, Sanghyun Woo, Manoj Middepogu,
  Sai~Charitha Akula, Jihan Yang, Shusheng Yang, Adithya Iyer, Xichen Pan,
  Ziteng Wang, Rob Fergus, Yann LeCun, and Saining Xie.
\newblock Cambrian-1: A fully open, vision-centric exploration of multimodal
  llms.
\newblock In \emph{Advances in Neural Information Processing Systems
  (NeurIPS)}, 2024.

\bibitem[Turner et~al.(2024)Turner, Thiergart, Leech, Udell, Vazquez, Mini, and
  MacDiarmid]{turner2024steeringlanguagemodelsactivation}
Alexander~Matt Turner, Lisa Thiergart, Gavin Leech, David Udell, Juan~J.
  Vazquez, Ulisse Mini, and Monte MacDiarmid.
\newblock Steering language models with activation engineering, 2024.
\newblock URL \url{https://arxiv.org/abs/2308.10248}.

\bibitem[Valentino et~al.(2026)Valentino, Kim, Dalal, Zhao, and
  Freitas]{valentino2026mitigatingcontenteffectsreasoning}
Marco Valentino, Geonhee Kim, Dhairya Dalal, Zhixue Zhao, and André Freitas.
\newblock Mitigating content effects on reasoning in language models through
  fine-grained activation steering, 2026.
\newblock URL \url{https://arxiv.org/abs/2505.12189}.

\bibitem[Wang et~al.(2024)Wang, Bai, Tan, Wang, Fan, Bai, Chen, Liu, Wang, Ge,
  Fan, Dang, Du, Ren, Men, Liu, Zhou, Zhou, and
  Lin]{wang2024qwen2vlenhancingvisionlanguagemodels}
Peng Wang, Shuai Bai, Sinan Tan, Shijie Wang, Zhihao Fan, Jinze Bai, Keqin
  Chen, Xuejing Liu, Jialin Wang, Wenbin Ge, Yang Fan, Kai Dang, Mengfei Du,
  Xuancheng Ren, Rui Men, Dayiheng Liu, Chang Zhou, Jingren Zhou, and Junyang
  Lin.
\newblock Qwen2-vl: Enhancing vision-language model's perception of the world
  at any resolution, 2024.
\newblock URL \url{https://arxiv.org/abs/2409.12191}.

\bibitem[Wang et~al.(2025)Wang, Jiao, Zhu, Chen, He, Chu, Gao, Wang, and
  Ma]{10.1145/3696410.3714640}
Tianlong Wang, Xianfeng Jiao, Yinghao Zhu, Zhongzhi Chen, Yifan He, Xu~Chu,
  Junyi Gao, Yasha Wang, and Liantao Ma.
\newblock Adaptive activation steering: A tuning-free llm truthfulness
  improvement method for diverse hallucinations categories.
\newblock In \emph{Proceedings of the ACM on Web Conference 2025}, WWW '25,
  pp.\  2562--2578, New York, NY, USA, 2025. Association for Computing
  Machinery.
\newblock ISBN 9798400712746.
\newblock \doi{10.1145/3696410.3714640}.
\newblock URL \url{https://doi.org/10.1145/3696410.3714640}.

\bibitem[Yang et~al.(2026)Yang, Cheng, Sun, Zhang, and
  Su]{10.1609/aaai.v40i40.40720}
Haocheng Yang, Xiang Cheng, Chenhao Sun, Pengfei Zhang, and Sen Su.
\newblock Privsv: differentially private steering vector for large language
  models.
\newblock In \emph{Proceedings of the Fortieth AAAI Conference on Artificial
  Intelligence and Thirty-Eighth Conference on Innovative Applications of
  Artificial Intelligence and Sixteenth Symposium on Educational Advances in
  Artificial Intelligence}, AAAI'26/IAAI'26/EAAI'26. AAAI Press, 2026.
\newblock ISBN 978-1-57735-906-7.
\newblock \doi{10.1609/aaai.v40i40.40720}.
\newblock URL \url{https://doi.org/10.1609/aaai.v40i40.40720}.

\bibitem[Zhang et~al.(2026)Zhang, Niu, Nan, Xu, and
  Weng]{10.1609/aaai.v40i41.40773}
Jie Zhang, Chenxu Niu, Zhefeng Nan, Yangyan Xu, and Jinta Weng.
\newblock Steering representations, safeguarding privacy: a cross-modal privacy
  protection method for generative ai.
\newblock In \emph{Proceedings of the Fortieth AAAI Conference on Artificial
  Intelligence and Thirty-Eighth Conference on Innovative Applications of
  Artificial Intelligence and Sixteenth Symposium on Educational Advances in
  Artificial Intelligence}, AAAI'26/IAAI'26/EAAI'26. AAAI Press, 2026.
\newblock ISBN 978-1-57735-906-7.
\newblock \doi{10.1609/aaai.v40i41.40773}.
\newblock URL \url{https://doi.org/10.1609/aaai.v40i41.40773}.

\bibitem[Zhang \& Sun(2025)Zhang and
  Sun]{zhang2025differentiateddirectionalinterventionframework}
Peng Zhang and Peijie Sun.
\newblock Differentiated directional intervention: A framework for evading llm
  safety alignment, 2025.
\newblock URL \url{https://arxiv.org/abs/2511.06852}.

\bibitem[Zou et~al.(2025)Zou, Phan, Chen, Campbell, Guo, Ren, Pan, Yin,
  Mazeika, Dombrowski, Goel, Li, Byun, Wang, Mallen, Basart, Koyejo, Song,
  Fredrikson, Kolter, and
  Hendrycks]{zou2025representationengineeringtopdownapproach}
Andy Zou, Long Phan, Sarah Chen, James Campbell, Phillip Guo, Richard Ren,
  Alexander Pan, Xuwang Yin, Mantas Mazeika, Ann-Kathrin Dombrowski, Shashwat
  Goel, Nathaniel Li, Michael~J. Byun, Zifan Wang, Alex Mallen, Steven Basart,
  Sanmi Koyejo, Dawn Song, Matt Fredrikson, J.~Zico Kolter, and Dan Hendrycks.
\newblock Representation engineering: A top-down approach to ai transparency,
  2025.
\newblock URL \url{https://arxiv.org/abs/2310.01405}.

\end{thebibliography}
